\documentclass[conference]{IEEEtran}
\IEEEoverridecommandlockouts
\usepackage{cite}
\usepackage{amsmath,amssymb,amsfonts}
\usepackage{algorithmic,algorithm}
\usepackage{graphicx}
\usepackage{textcomp}
\usepackage{xcolor}
\usepackage{float}
\usepackage{multirow}

\allowdisplaybreaks[4]

\def\BibTeX{{\rm B\kern-.05em{\sc i\kern-.025em b}\kern-.08em
    T\kern-.1667em\lower.7ex\hbox{E}\kern-.125emX}}
    
\begin{document}

\title{
Semi-Asynchronous Federated Edge Learning for Over-the-air Computation
% FedAPA: A Semi-asynchronous Federated Learning Mechanism via Wireless Multiple-Access Channels
% Power Control for Asynchronous Federated Learning via Wireless Multiple-Access Channels
}

\author{\IEEEauthorblockN{Zhoubin Kou\textsuperscript{1},\
Yun Ji\textsuperscript{1},
Xiaoxiong Zhong\textsuperscript{1,2}, and 
Sheng Zhang\textsuperscript{1,*}}
\IEEEauthorblockA{\textsuperscript{1}Graduate School in Shenzhen, Tsinghua University, Shenzhen, 518055, China}
\IEEEauthorblockA{\textsuperscript{2}Peng Cheng Laboratory, Shenzhen 518000, P.R. China}
\IEEEauthorblockA{\textsuperscript{*}Corresponding author: Sheng Zhang, email: zhangsh@sz.tsinghua.edu.cn}}

\maketitle
\begin{abstract}
    % As a machine learning manner where many edge devices collaboratively train a model without data sharing, federated learning (FL) has been investigated extensively under wireless transmission scenarios. However, the traditional synchronous FL mechanism in wireless communications suffers from stragglers issues and limited resources. In this paper, we propose a semi-asynchronous FL mechanism following the periodic aggregation mechanism (FedAPA) via the wireless multi-access channels and derive the convergence upper bound of this mechanism. Taking the staleness and divergence of model updates into consideration, we optimize the convergence behavior of the FL global model by adjusting the uplink transmit power of edge devices at each round. The simulation results reveal that our proposed FedAPA algorithm can achieve faster and higher convergence speed than the synchronous FL, full asynchronous FL, and time-triggered FL in the highly non-IID settings.
    
    Over-the-air Computation (AirComp) has been demonstrated as an effective transmission scheme to boost the efficiency of federated edge learning (FEEL). However, existing FEEL systems with AirComp scheme often employ traditional synchronous aggregation mechanisms for local model aggregation in each global round, which suffer from the stragglers issues.
    In this paper, we propose a semi-asynchronous aggregation FEEL mechanism with AirComp scheme (PAOTA) to improve the training efficiency of the FEEL system in the case of significant heterogeneity in data and devices.
    Taking the staleness and divergence of model updates from edge devices into consideration, we minimize the convergence upper bound of the FEEL global model by adjusting the uplink transmit power of edge devices at each aggregation period.
    The simulation results demonstrate that our proposed algorithm achieves convergence performance close to that of the ideal Local SGD. Furthermore, with the same target accuracy, the training time required for PAOTA is less than that of the ideal Local SGD and the synchronous FEEL algorithm via AirComp.
    
    \end{abstract}
    
\begin{IEEEkeywords}
    Federated edge learning, semi-asynchronous mechanism, over-the-air computation.
    \end{IEEEkeywords}
% \clearpage
\section{introduction}
% --------------------------------------
% 第一块：引入联邦学习，为啥需要联邦学习
% --------------------------------------
With the advancement in computing capabilities and the accessibility of an unprecedented amount of data for portable devices, various machine learning based applications and services has been introduced to Internet of Thing (IoT) systems. 
However, the frequent data sharing of individual information in some services led to privacy concerns.
Due to its appealing features of privacy protection, federated learning (FL) has been widely regarded as a promising machine learning technology \cite{b1}.
Nevertheless, there are several problems to be addressed when deploying FL on wireless scenarios: 
1) resource limitation: the total bandwidth and the transmission energy for all edge devices in wireless FL system are both finite;
2) heterogeneity: wireless FL suffers from the data heterogeneity and device heterogeneity, leading to global non-IID data distribution and different computing latency respectively. 

% --------------------------------------
% 第二块：联邦学习使用多址实现训练的存在的问题
% --------------------------------------
In federated edge learning (FEEL) scenario, edge devices that collaborate to build a global model are often dispersed within a small physical range, and coordinated by a nearby parameter server (PS) \cite{b2}. 
% And how to upload local models efficiently from edge devices to the BS via wireless transmission is a critical challenge for achieving effective federated learning. 
A traditional approach for the uplink transmission of edge devices is to allocate channel resources through orthogonal access techniques, such as TDMA, CDMA and OFDMA.
% However, to ensure the reliability of communication between edge devices and the BS, there is an upper limit on the number of edge devices that can participate in uploading under limited wireless communication resources
% However, considering the limited wireless resources and the large size of model data transmitted in FEEL, there exists an upper limit on the number of edge devices that can participate in uploading while ensuring the reliability of communication between edge devices and the BS.
However, the limited wireless resources and large model data size in FEEL impose a constraint on the number of edge devices that can participate in uploading. 
% As a distributed learning scheme, FEEL allows the edge devices to build a global model collaboratively while preserving the local training data on their own equipment, such as FedAvg proposed in \cite{b1}. 

% --------------------------------------
% 第三块：OTA在FEEL中的应用
% --------------------------------------
Over-the-air computation (AirComp) has been proven to be an efficient paradigm to alleviate the communication costs and accelerate the FEEL training progress \cite{T2020OTA}. Leveraging the superposition property of wireless spectrum, AirComp can achieve uplink transmission of local models without the need for spectrum and time channel resource allocation.

To enhance the performance of wireless FEEL system, several studies related to the AirComp system for FEEL have been done \cite{T2020OTA, OTA_misalignment, Yang2020OTA}. A large amount of works follow the synchronous aggregation FEEL mechanism , where the parameter server (PS) does not update the global model until it receives local models from all the chosen edge devices in each global aggregation. 
However, in the common FEEL scenario of imbalanced computing ability among edge devices, following the synchronous aggregation mechanism for training FEEL can result in the risk of bottleneck nodes, while discarding the stragglers may need to lose important data.
The former can prolong the training time of the model, and the latter can decrease the predictive accuracy of the final model.

% However, since the parameter server (PS) does not update the global model until it receives local models from all the chosen edge devices in each global round, there is a risk of latency caused by the stragglers whose local model carried important information to global training.

% --------------------------------------
% 第四块：异步无线联邦的研究现状
% --------------------------------------
In these circumstances, one potential solution is to apply the asynchronous mechanism into the FEEL. To fully take the advantage of high uplink throughput in AirComp, we propose a semi-asynchronous model aggregation mechanism with fixed interval time for global aggregation. 
% In \cite{b5}, the author designed a asynchronous FEEL framework using adaptive parameter to improve its scalability and FEELexibility, and a central model fusion algorithm was proposed to balance the inFEELuence of staleness raised by asynchronous mechanism.
% In \cite{TT-Fed}, a time-triggered FEEL algorithm over wireless systems was proposed to train FEEL problems, and the training loss was minimized by joint device selection and bandwidth optimization.
% These studies both consider the uplink transmission as frequency division multiple access (FDMA) technique, which means the upload conFEELicts may occur because of the limited bandwidth resources.
% % --------------------------------------
% % 第五块：异步无线联邦+over the air(提出本文的方案)
% % --------------------------------------
% In this paper, we propose a novel semi-asynchronous FEEL framework via wireless multiple-access (MAC) channels. Because of the waveform superposition nature of the MAC channels, the edge devices can share the same bandwidth resource at each upload time slot and transmit the model information by over-the-air computation (AirComp) \cite{over-the-air}.
% In this way, the stragglers issue introduced by the synchronous mechanism can be mitigated as well as the increased complexity caused by congestion controlling can be prevented. 
The main contributions are summarized as follows:
\begin{itemize}
\item We propose a semi-asynchronous \underline{P}eriodic \underline{A}ggregation \underline{O}ver-\underline{T}he-\underline{A}ir computation strategy named PAOTA under wireless multiple access channel (MAC) scenarios, which can utilize the waveform superposition property to realize AirComp during the wirless transmission.
% With respect to the PAOTA, we derive the convergence upper bound, and analyze the influence of asynchronous mechanism and wireless environment on the convergence of the algorithm. 
\item We analyze the convergence behavior of the semi-asynchronous FEEL via AirComp and derive the upper bound of the gap between the expected and optimal global loss values with respect to the transmission power.
In PAOTA, the weighted parameters for model aggregation are proportional to the transmission power of the edge devices.
To overcome the staleness in asynchronous aggregation and alleviate the data skewness caused by non-IID data, we transform the power control for uplink transmission into a trade-off optimization for the delay factor of local model and the similarity factor of model gradient.
\item The trade-off optimization problem is a nonlinear fractional programming of two convex quadratic functions. We solve it by using Dinkelbach's parametrization scheme. To optimize the nonconcave quadratic programming problem introduced by Dinkelbach's transform, we convert the  problem into a 0-1 linear integer programming problem through piecewise linear approximation.
% Combining the characteristics of asynchronous updates, we transform the power control for uplink transmission into a trade-off optimization for the delay factor of local model and the similarity factor of model gradient.
\item According to the numerical results, PAOTA shows good training robustness under wireless FEEL system. Considering the heterogeneity of FEEL in the experiments, we verify the superiority of the algorithm PAOTA in terms of the predictive accuracy and the training time for converging to the target prediction accuracy. 
% and the performance of our algorithms is verified by extensive simulations.
\end{itemize}
% \clearpage
\section{System Model}
% --------------------------------------
% 第一块：介绍联邦学习的优化目标（优化问题的公式）
% --------------------------------------

\subsection{FL Problems}
\begin{figure}[t]
\centerline{\includegraphics[width=0.5\textwidth]{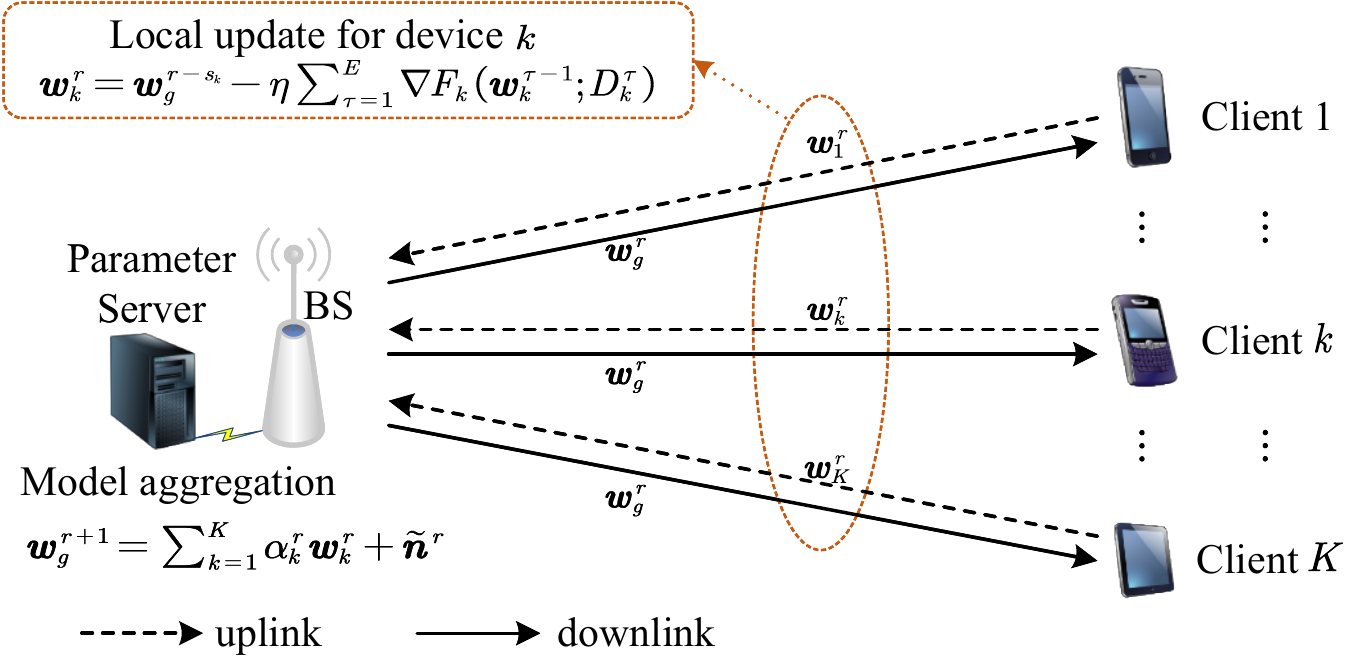}}
\caption{FL system via wireless multiple-access channels.}
\label{fig:sys_mac}
% \vspace{-0.3cm}
\end{figure}

We consider a wireless FL system which consists of one parameter server and a set $\mathcal{K}=\left\{1,\dots,K\right\}$ of $K$ edge devices, as shown in Fig.~\ref{fig:sys_mac}. 
Each client $k\in\mathcal{K}$ participating in the FL task is access to a local data set $\mathcal{D} _k=\left\{ \left( \boldsymbol{x}_{k,1},y_{k,1} \right) ,\dots ,\left( \boldsymbol{x}_{k,D_k},y_{k,D_k} \right) \right\} $, with size $\vert \mathcal{D}_k\vert = D_k$.
Then, the total set of data samples in the whole system can be denoted as $\mathcal{D}=\left\{\mathcal{D}_1,\dots,\mathcal{D}_K\right\}$, where size $D=\sum\nolimits_{k=1}^K D_k$.
$\left( \boldsymbol{x}_{k,i},y_{k,i} \right)$ is the $i$-th input-output pair stored in client $k$, where $\boldsymbol{x}_{k,i}$ denotes the feature vector and $y_{k,i}$ denotes the corresponding label value.
%----- fig1: 模型聚合机制图片
The goal of FL task is to minimize the global loss function by training global model parameter $\boldsymbol{w}$, where the local data samples at edge devices are unavailable to the PS on account of the privacy concern.
The optimization problem of FL can be formulated as follow:
\begin{equation}\label{Eq:sys_FL_task}
    \min_{\boldsymbol{w}}\ F(\boldsymbol{w})=\sum\nolimits_{k=1}^K \frac{D_k}{D}F_k(\boldsymbol{w}),
\end{equation}
where $F_k$ is the local loss function defined as 
\begin{equation}\label{Eq:sys_local_loss}
    F_k(\boldsymbol{w})=\frac{1}{D_k}\sum\nolimits_{i\in D_k}l(\boldsymbol{w}; \left( \boldsymbol{x}_{k,i},y_{k,i} \right)),
\end{equation}
where $l(\boldsymbol{w}; \left( \boldsymbol{x}_{k,i},y_{k,i} \right))$ is the empirical loss function.

\subsection{Semi-asynchronous FL with Periodic Aggregation}
% --------------------------------------
% 第二块：介绍联邦学习的聚合机制
% --------------------------------------
Inspired by \cite{TT-Fed}, we propose a time-triggered semi-asynchronous aggregation.
The work-flow of our proposed PAOTA is shown in Fig.~\ref{fig:sys_aggregation}.
The global aggregation proceeds periodically, and the interval time of each cycle $\Delta T$ remains constant.
We assume $R$ rounds of FL training are performed, and use vector $\boldsymbol{b}^r=[b_1^r,\cdots,b_K^r]\in\{0,1\}$ to indicate the state information of $K$ edge devices. 
% At each round, the elements of state vector are assigned as zero at first. 
After the PS broadcasts the model to client $k$, the element of state vector $b_k^r$ is assigned as zero at first. 
When client $k$ finishes its local training at the $r$-th round, it sends a signal to the server showing that client $k$ is ready to upload its local model.
Then the PS will set the value of $b_k^r$ to $1$, which means client k will attend the global aggregation at $r$-th round and be ready to receive the updated global model at the beginning of the  $(r$$+$$1)$-th round. 
% From a system scheduling perspective, our proposed semi-asynchronous aggregation mechanism can also be referred to as a time-triggered asynchronous approach. 
% In \cite{TT-Fed}, the authors designed a tire-based time-triggered federated learning with OFDMA for uplink transmission. And our proposed mechanism can be seen as time-triggered FEEL with only one tire, because of AirComp does not require the users to divide the channel resources among each other and communicate at higher throughput uplink communications.
\subsection{Asynchronous Aggregation via AirComp}
% 无线通信模型设计，上行链路、下行链路的设置
In this paper, we consider the scenario where the downlink communication is error-free. 
The uplink channels remain unchanged when edge devices transmit local models to the PS at one round, and the wireless MAC channels are adopted. 
Following the Rayleigh distribution, the uplink channels are independent across different transmission rounds.

%----- fig: 模型聚合机制图片
\begin{figure}[t]
\centerline{\includegraphics[width=0.5\textwidth]{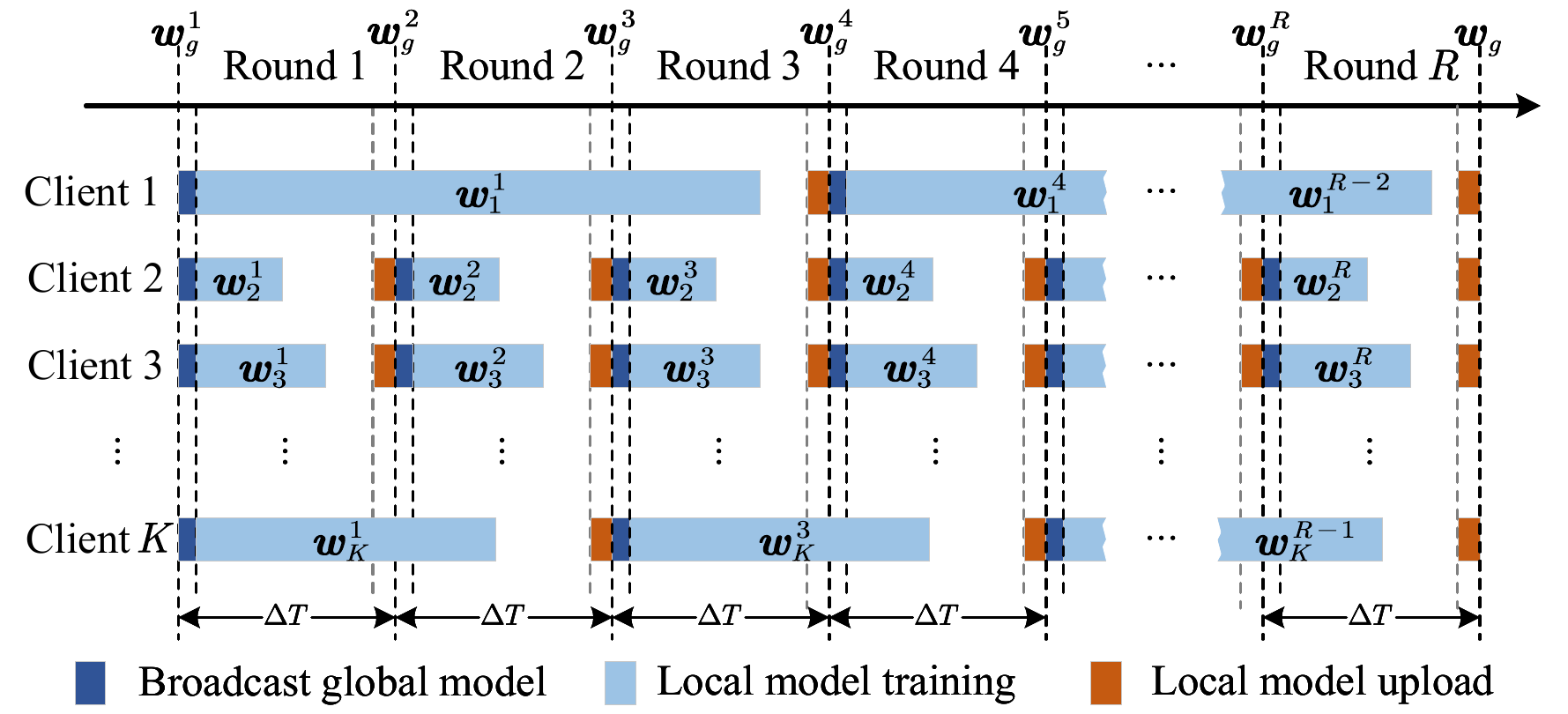}}
\caption{The work-flow of federated learning with periodic aggregation.}
\label{fig:sys_aggregation}
% \vspace{-0.3cm}
\end{figure}

% 异步联邦学习的实现机制，包含相关的公式介绍
We assume the PS will carry out $R$ rounds of iterations in total for a FL task.
At the $r$-th ($r\in\{1,\cdots, R\}$) round of global training iteration:

(a) \textbf{Global model broadcasting}. At the beginning of each global round, the PS broadcasts the global model $\boldsymbol{w}_\text{g}^k$ to edge devices according to the edge devices state vector $\boldsymbol{b}^r=[b_1^r,\cdots,b_K^r]\in\{0,1\}$.
When client $k$ does not complete the local training of the previous global round, set $b_k^r=0$; and when client $k$ is ready to participate in the FL task at the $r$-th round, $b_k^r=1$. 
In particular, the PS broadcasts global model $\boldsymbol{w}_{\text{g}}^1$ to all edge devices at the first global round, which means ${b}^1_k=1$, $\forall k$.
Since the downlink communication is assumed to be a perfect transmission, each client can get the global model without transmission errors. 

(b) \textbf{Local model training}. 
If $b_k^r=1$, client $k$ receives the global model $\boldsymbol{w}_{\text{g}}^r$ and use  stochastic gradient decent (SGD) to update local model based ont their own datasets $\mathcal{D}_k$.
% updates local model, based on the local dataset , to minimize the local loss function $\eqref{Eq:sys_local_loss}$ via stochastic gradient decent (SGD) algorithm. 
We assume each client runs $M$ rounds iterations for local model training, at the $m$-th local update, $m\in\{1,\cdots,M\}$, the local model $\boldsymbol{w}_{k,m}^{r}$ is formed as
\begin{equation}\label{Eq:local_update}
    \boldsymbol{w}_{k,m}^{r}=\boldsymbol{w}_{\text{g}}^{r}-\eta\sum\nolimits_{\tau=1}^m\nabla F_k(\boldsymbol{w}_{k,\tau-1}^{r};D_k^{\tau}),
\end{equation}
where $\eta$ is the learning rate, $D_k^m$ is the data used in the $m$-th round of local training.
If $b_k^r=0$, client $k$ will keep training the local model which haven't been finished at the previous round. We define $s_k^r$ as the number of rounds that client $k$ falls behind the global training round, then the update of straggler $k$ can be expressed as 
\begin{equation}\label{Eq:local_update_stagger}
    \boldsymbol{w}_{k,m}^{r,s_k^r}=\boldsymbol{w}_{\text{g}}^{r-s_k^r}-\eta\sum\nolimits_{m=1}^M\nabla F_k(\boldsymbol{w}_{k,m-1}^{r,s_k^r};D_k^{m}),
\end{equation}

As we can see, \eqref{Eq:local_update} is a special case of \eqref{Eq:local_update_stagger} when $s_k^r$ equals to 0.
After $M$ SGD training iterations, client $k$ finishes its local training at $r$-th global round and the updated local model is denoted as $\boldsymbol{w}_{k,M}^{r,s_k^r}$. 

(c) \textbf{Local model upload.} In this paper, we assume that the wireless AirComp can achieve strict clock synchronization, called alignment over-the-air computation, so that edge devices' signals overlap exactly with each other at the PS.

% When edge devices finish their local training, they can transmit the updated local models to the PS at the upload time slot of this round. 
The edge devices who complete their local training during $r$-th global round will transmit their local model through AirComp at the same period, as shown in Fig. \ref{fig:sys_aggregation}.
Assuming the channel state information (CSI) is known by all the edge devices and the PS perfectly, we consider a pre-processing parameter $\phi_k^r$ for the transmitter of the edge devices which is denoted by
\begin{equation}\label{Eq:pre-processing_parameter}
    \phi_k^r=\frac{b_k^r p_k^r (h_k^r)^H}{\vert h_k^r\vert^2},
\end{equation}
where $h_k^r\in\mathbb C$ is the complex channel coefficient of the uplink transmission between client $k$ and the PS, $p_k^r$ is the transmit power of client $k$, and $(\cdot)^H$ represents conjugate transpose. 

The transmit signal of client $k$ can be expressed as $\boldsymbol{x}_k^r=\phi_k^r\boldsymbol{w}_{k,M}^{r,s_k^r}$.
As the PS knows the perfect CSI, the received signal of PS can be formulated as
\begin{equation}\label{Eq:Global_receive}
    \begin{aligned}
        \boldsymbol{y}_{\text{g}}^r=\sum\nolimits_{k=1}^K h_k^r \boldsymbol{x}_k^r + \boldsymbol{n}^r=\sum\nolimits_{k=1}^K b_k^r p_k^r \boldsymbol{w}_{k,M}^{r,s_k^r} + \boldsymbol{n}^r,
    \end{aligned}
\end{equation}
where $\boldsymbol{n}^r\in\mathbb{C}^d$ represents the independent identical distribution (i.i.d) additive Gaussian white noise (AWGN) following the distribution $\mathcal{CN}(0, \sigma_n^2\boldsymbol{I})$. And $\sigma_n^2=BN_0$, where $B$ is the bandwidth of uplink channel, and $N_0$ represents the channel noise power spectral density.
% Considering there is limited power in every client, the transmit power of each client is supposed to satisfy the following constraint:
For limited power of each client, we have:
\begin{equation}\label{Eq:individual_power}
    \Vert \phi_k^r\boldsymbol{w}_{k,M}^{r,s_k^r}\Vert_2^2 \leq P_{k,\text{max}}^r,
\end{equation}
where $P_{k,\text{max}}^r$ is the maximum power of client $k$ at the $r$-th round.

% As shown in Fig. \ref{fig:sys_aggregation}, edge devices that finish local training will simultaneously upload local models during the same time period, ensuring the realization of AirComp.
(d) \textbf{Global model update}. After the upload time slot of each global round, the PS receives the aggregation signal through AirComp, and performs a normalization operation to obtain the updated global model $\boldsymbol{w}_{\text{g}}^{r+1}$ 
\begin{equation}\label{Eq:Normalization}
    \begin{aligned}
         \boldsymbol{w}_{\text{g}}^{r+1}&=\frac{\boldsymbol{y}_{\text{g}}^r}{\varsigma^r}= \sum\nolimits_{k=1}^K \frac{b_k^r p_k^r}{\varsigma^r} \boldsymbol{w}_{k,M}^{r,s_k^r} + \frac{\boldsymbol{n}^r}{\varsigma^r}\\
         &= \sum\nolimits_{k=1}^K \alpha^r_k \boldsymbol{w}_{k,M}^{r,s_k^r} + \tilde{\boldsymbol{n}}^r,
    \end{aligned}
\end{equation}
where $\varsigma^r$ is the normalization factor at $r$-th round and can be calculated by $\varsigma^r=\sum\nolimits_{k=1}^K b_k^r p_k^r$. Then the actual weight parameter of client $k$ can be formulated by $\alpha_k^r=\tfrac{b_k^r p_k^r}{\sum\nolimits_{i=1}^K b_i^r p_i^r}$  for simplicity, where $\alpha_k^r$ satisfies $\sum\nolimits_{k=1}^K \alpha_k^r=1$. $\tilde{\boldsymbol{n}}^r$ is the equivalent noise after the normalization operation.

To facilitate the derivation of convergence, we further express global model aggregation as follow:
\begin{equation}\label{aggre-final}
    \begin{aligned}
    \boldsymbol{w}_\text{g}^{r+1}
    % =& \sum\nolimits_{k=1}^K \alpha^r_k \boldsymbol{w}_{\text{g}}^{r-s_k^r}
    % &-\eta\sum\nolimits_{k=1}^K\sum\nolimits_{\tau=1}^E\nabla F_k(\boldsymbol{w}_{k,\tau-1}^{r,s_k^r};D_k^{\tau}) \\\\
    % & + \tilde{\boldsymbol{n}}^r \\
    = \tilde{\boldsymbol{w}}^r + \sum\nolimits_{k=1}^K\alpha_k^r\Delta\boldsymbol{w}_k^r + \tilde{\boldsymbol{n}}^r,
    \end{aligned}
\end{equation}
where $\tilde{\boldsymbol{w}}^r=\sum\nolimits_{k=1}^K \alpha^r_k \boldsymbol{w}_{\text{g}}^{r-s_k^r}$ represents the weighted sum of the global model parameters based on the users participating in uploading the local model during $r$-th round of aggregation, and $\Delta\boldsymbol{w}_k^r=-\eta\sum\nolimits_{\tau=1}^E\nabla F_k(\boldsymbol{w}_{k,\tau-1}^{r,s_k^r};D_k^{\tau})$ is the local update of client $k$ at the $r$-th round.

% \clearpage
\section{Convergence Analysis and Optimization Algorithm}
% As mentioned before, we propose a semi-asynchronous federated learning algorithm with periodic aggregation via the wireless MAC channels.
% What we are interested in is how wireless environment and the asynchronous aggregation strategy affect the convergence speed and accuracy of the algorithm we proposed. 
% At first, we derive the upper bound of the expected optimal gap between the expected and optimal global loss values. Then, combining with the characteristics of the asynchronous mechanism, we minimize the derived bound by optimizing the parameter $\alpha$ related to the staleness factor and the interference factor.
In this section, we analyze how the wireless MAC environment and the periodic aggregation strategy affect the convergence behavior of PAOTA presented in Section II. 
Firstly, we derive the upper bound of the expected optimal gap between the expected and optimal global loss values. 
Then, combining with the characteristics of the asynchronous mechanism and data heterogeneity, we minimize the derived upper bound by optimizing the parameter $\boldsymbol{\beta}$ related to the staleness factor and the gradient similarity factor. The whole process of PAOTA is shown in \textbf{Algorithm 1}.

\subsection{Convergence Analysis}
We present the following assumptions and lemmas that are necessary when we derive the convergence behavior of PAOTA algorithm at first.

\emph{\textbf{Assumption 1}: The global loss function $F$ is $L$-smooth, \emph{i.e.}, $\forall \boldsymbol{x}, \boldsymbol{y}$:}
\begin{equation}\label{ass1-1}
    F(\boldsymbol{x})-F(\boldsymbol{y})\leq (\boldsymbol{x}-\boldsymbol{y})^T\nabla F(\boldsymbol{y})+\frac{L}{2}\Vert \boldsymbol{x}-\boldsymbol{y}\Vert^2_2,
\end{equation}
\begin{equation}
    \Vert \nabla F(\boldsymbol{x})-\nabla F(\boldsymbol{y})\Vert\leq L\Vert \boldsymbol{x}-\boldsymbol{y}\Vert,\label{ass1-2}
\end{equation}

\emph{\textbf{Assumption 2} \cite{over-the-air}: The variance of the local model gradients at each local device is bounded by} $\zeta$:
\begin{equation}\label{ass2-1}
    \mathbb{E} \left[ \left\| \nabla F_k\left( \boldsymbol{w}_{k}^{\tau}\right) -\nabla F\left( \boldsymbol{w}_\text{g} \right) \right\| _{2}^{2} \right] \le \zeta,
\end{equation}
\emph{where $\zeta$ is the parameter related to the data heterogeneity.}

These two assumptions above are widely used in the convergence analysis for traditional synchronous FL. 
\emph{\textbf{Assumption 1}} makes sure the gradient of $F$ does not change quickly during global training. 
And the \emph{\textbf{Assumption 2}} captures the degree of data heterogeneity by parameter $\zeta$.

\emph{\textbf{Assumption 3} \cite{TT-Fed}: The global model gradient change within $n$ training rounds is bounded as}
\begin{equation}\label{ass3-1}
    \left( \boldsymbol{w}_\text{g}^{r-n} -\boldsymbol{w}_\text{g}^r  \right) ^T\nabla F\left( \boldsymbol{w}_\text{g}^r  \right) \le \delta \left\| \nabla F\left( \boldsymbol{w}_\text{g}^r  \right) \right\| ^2_2,
\end{equation}
\begin{equation}\label{ass3-2}
    \Vert\boldsymbol{w}_\text{g}^{r-n} -\boldsymbol{w}_\text{g}^r\Vert\leq\epsilon,
\end{equation}
\emph{where $\delta$ and $\epsilon$ are constant value. And the local model gradient change within $m$ local rounds is bounded as}
\begin{equation}\label{ass3-3}
    \Vert \nabla F(\boldsymbol{w}_k^{r-m})\Vert\leq\vartheta\Vert\nabla F(\boldsymbol{w}_k^{r})\Vert,
\end{equation}
\emph{where $\vartheta$ is a constant value.}

% The above assumption can be satisfied by the widely used loss function. It ensures that the global update and local computation both do not change so fast that the stale local models updated by the clients can still provide useful information for global model training. 
 
\emph{\textbf{Assumption 4} : The SGD algorithm performed by the edge device is unbiased, i.e.,}
\begin{equation}
    \mathbb{E}[\nabla F_k(\boldsymbol{w}_k;\mathcal{D}_k)]=\nabla F_k(\boldsymbol{w}_k),
\end{equation}
\emph{and the variance of stochastic gradients at each edge device is bounded as}
\begin{equation}
    \mathbb{E}[\Vert \nabla F_k(\boldsymbol{w}_k;\mathcal{D}_k)-\nabla F_k(\boldsymbol{w}_k)\Vert_2^2]\leq \sigma^2,
\end{equation}
\emph{where $\sigma$ is a constant value.}

\emph{\textbf{Lemma 1}: The sum of the expected square norm of the difference between the local updated model at each SGD iteration and the previous global model is bounded by }
\begin{equation}\label{lemma1}
    \begin{aligned}
    & \sum\nolimits_{\tau=1}^M\mathbb{E}[\Vert\boldsymbol{w}_\text{g}^{r-s_k^r}-\boldsymbol{w}_k^{r-s_k^r,\tau-1}\Vert_2^2]\\
    \leq & \frac{{\eta _t}^2M^3\sigma ^2+4{\eta _t}^2M^3L^2\zeta +4{\eta _t}^2M^3\beta ^2\mathbb{E} \left[ \left\| \nabla F\left( \boldsymbol{w}_\text{g}^r \right) \right\| _{2}^{2} \right]}{1-2{\eta}^2M^2L^2},
    \end{aligned}
\end{equation}

\begin{IEEEproof}
See the section Appendix A.
\end{IEEEproof}

\emph{\textbf{Lemma 2}: For a $L$-smooth function $F$ with optimum solution $\boldsymbol{w}^*$, the following inequality holds}
\begin{equation}\label{eq:lemma2}
    \Vert\nabla F(\boldsymbol{w})\Vert_2^2\leq 2L(F(\boldsymbol{w})-F(\boldsymbol{w}^*)),
\end{equation}
\begin{IEEEproof}
As $F$ is a $L$-smooth function, for $\boldsymbol{w}$ and the optimum solution $\boldsymbol{w}^*$, we have
\begin{equation}\label{eq:proof2}
\begin{aligned}
    &\tfrac{1}{2L}\Vert \nabla F(\boldsymbol{w})-\nabla F(\boldsymbol{w}^*)\Vert_2^2\\
    \leq & F(\boldsymbol{w})-F(\boldsymbol{w}^*)-(\boldsymbol{w}-\boldsymbol{w}^*)^T\nabla F(\boldsymbol{w}^*),
\end{aligned}    
\end{equation}
where $\nabla F(\boldsymbol{w}^*)=0$. Rearrange the \eqref{eq:proof2} and we can get the \eqref{eq:lemma2}.
\end{IEEEproof}
Now, we present the main convergence analysis result in the following theorem.
Now, we present the main convergence analysis result in the following theorem.

\emph{\textbf{Theorem 1}: The expected optimal gap between the expected and optimal global loss values is bounded as }
\begin{equation}
    \begin{aligned}
    &\mathbb{E}[F(\boldsymbol{w}^{R+1})]-F(\boldsymbol{w}^*)\\ \leq&
    \prod_{r=1}^R{A^r\mathbb{E} \left[ F\left( \boldsymbol{w}_\text{g}^1 \right) -F(\boldsymbol{w}^*) \right]} +G^R\\
    &+\sum_{r=1}^R{\left( \prod_{i=r+1}^R{A^i} \right) G^r },
    \end{aligned}
\end{equation}
\emph{where} 
\begin{equation}
\begin{aligned}
    A^r = &1+2L\delta -L\eta M+8L^2\eta^{2}M\vartheta ^2\\
    &+\left( \eta L^2+4M\eta^{2}L^3 \right) \frac{8L\eta^{2}M^3\vartheta ^2}{1-2\eta^{2}M^2L^2},
\end{aligned}
\end{equation}
\emph{and}
\begin{equation}\label{eq:G^r}
\begin{aligned}
    G^r =&\underbrace{(2\eta M+8L\eta M^2+\frac{4\eta^2M^3L^2(\eta L^2+4M\eta^2L^3)}{1-2\eta^2M^2L^2})\zeta}_{(a)}+ \\
    & \underbrace{2\eta ML^2\epsilon^2}_{(b)}+\underbrace{(2\eta^2 LM^2+\frac{(\eta L^2+4M\eta^2L^3)\eta^2M^3}{1-2\eta^2M^2L^2})\sigma^2}_{(c)}\\
    &+ \underbrace{L\epsilon^2K\sum\nolimits_{k=1}^K(\alpha_k^r)^2}_{(d)}+ \underbrace{\frac{2Ld\sigma_n^2}{(\sum\nolimits_{k=1}^K b_k^r p_k^r)^2}}_{(e)},
\end{aligned}
\end{equation}
\begin{IEEEproof}
Due to space limitations, please see Appendix A in the extended version \cite{arxiv}.
\end{IEEEproof}

According to the \emph{\textbf{Theorem 1}}, we can learn that the upper bound of $\mathbb{E}[F(\boldsymbol{w}^{R+1})]-F(\boldsymbol{w}^*)$ depends only on the second term $G^r$ given a sufficient number of iteration rounds, as long as the setting of learning rate $\eta$ satisfies $A(t)<1$. 

It is natural to think of minimizing the value of $G^r$ by adjusting the controllable parameters in the wireless FL system.
As shown in \eqref{eq:G^r}, $G^r$ consists of 5 terms $(a)$-$(e)$:
the terms $(a)$-$(c)$ are only dependent on the hyper-parameters that relate to the wireless FL system settings, which can not change during the training iterations.
Term $(d)$ and term $(e)$ contain the upload transmit power value $p_k$, which control the aggregation weight of the local models uploaded by different edge devices.

To sum up, there are two kinds of factors affecting the global model convergence in our proposed system. On the one hand, the asynchronous aggregation process introduces stale models to global update, thus impairing the convergence speed of FEEL. On the other hand, the noise present in the wireless transmission environment negatively impacts the convergence performance of federated learning.

\renewcommand{\algorithmicrequire}{ \textbf{Require:}} %Use Input in the format of Algorithm
\renewcommand{\algorithmicensure}{ \textbf{Initialize:}} %Use Output in the format of Algorithm
\begin{algorithm}[t]
\caption{PAOTA}
\label{alg:PAOTA}
\begin{algorithmic}[1] %这个1 表示每一行都显示数字
\REQUIRE  %算法的输入参数：Input
    Global training rounds $R$; Time duration of each round $\Delta T$; Local training rounds $E$; Uplink power budget of each client $p^{\text{max}}_k$; Initial global model $\boldsymbol{w}_\text{g}^0$; State Tag of each client $b_k^0=1$,$\forall k$;\\
    \FOR{$r=0,1, \cdots, R-1$}
        \STATE PS broadcasts $\boldsymbol{w}_\text{g}^r$ to clients $k$ satisfying $b_k=1$, $\forall k$;
        \STATE Set $b_k\leftarrow 0$;
        \FOR{$k=1, \cdots, K\ \text{in parallel}$ }
            \FOR{$\tau=1,\cdots, M$}
                \STATE $\boldsymbol{w}_k^{r,\tau+1}\leftarrow\boldsymbol{w}_k^{r,\tau}-\eta\nabla F_k(\boldsymbol{w}_k^{r,\tau};D_k^{\tau+1})$;
                \IF{$\tau=M$}
                    \STATE $b_k\leftarrow 1$;
                    \STATE Obtain current round value $r'$, uplink channel gains $h_k^{r'}$ and parameter $\boldsymbol{\beta}^{r'}$;
                    \STATE Set the $\phi_k^{r'}$ based on \eqref{Eq:pre-processing_parameter} and \eqref{asy:power_control};
                    \STATE Transmit the signal $\boldsymbol{x}_k^{r', r'-r}=\phi_k^{r'}\boldsymbol{w}_{k, E}^r$ at the aggregation time slot of round $r'$;
                \ENDIF
            \ENDFOR 
        \ENDFOR
        \STATE PS receives MAC signal \eqref{Eq:Global_receive} at the aggregation time slot, performs the normalization operation based on \eqref{Eq:Normalization} and obtains the updated global model $\boldsymbol{w}_\text{g}^{r+1}$;
    \ENDFOR
    % \vspace{-0.3cm}
\end{algorithmic}
\end{algorithm}
% term $(a)$ is caused by data heterogeneity $\zeta$; term $(b)$ is introduced by the asynchronous mechanism; term $(c)$ is caused by the data partitioning setting for SGD algorithm; term $(d)$ is affected by the asynchronous mechanism and the uplink transmit power of each client; and term $(e)$ is caused by the AWGN in the uplink channel and the uplink transmit power of each client.
% The terms $(a)$-$(c)$ are depended on the hyper-parameters that cannot change during the training iterations. 

\subsection{Power Control Optimization}
Based on the \emph{\textbf{Theorem 1}}, we can minimize the upper bound of $\mathbb{E}[F(\boldsymbol{w}^{R+1})]-F(\boldsymbol{w}^*)$ by optimizing the terms $(d)$ and $(e)$ through the uplink transmit power $p_k$, $k=1,\cdots, K$. By dropping the notation $r$ for simplicity, the optimal problem can be formulated as:
\begin{subequations}\label{opt1:problem}
    \begin{align}	   
        \mathbf{P1}:\ \ {\min_{p_1,\cdots,p_K}} \ \ & L\epsilon ^2K\sum\nolimits_{k=1}^K{{\alpha_k}^2}+\frac{2Ld\sigma _{n}^{2}}{\left( \sum\nolimits_{k\in \mathcal K}{b_kp_k} \right) ^2}\label{opt1:objective}\\
        {\text{s.t.}}\ \ &  p_k\le P_{\text{max}}^k,k=1,\cdots,K, \label{opt1:a}
        % & \boldsymbol{p}\in \mathcal{P} \label{opt1:p}
        % & p_k^2q_k\le P_{\text{max}}^k,k=1,\cdots,K, \label{opt1:p_k}
    \end{align}
\end{subequations}    
where $\boldsymbol{p}=\left[p_1,\cdots,p_K\right]\in R^{1\times K}$ is the transmission power of $K$ clients. 

Different with synchronous FL, PAOTA has to suffer the impact of the stale information. Meanwhile, the problem of data bias introduced by the non-IID data distribution also should be considered. As the model aggregation weights is determined by the uplink transmission power directly according to \eqref{Eq:Normalization}, we represent the power parameter as follow:
\begin{equation}\label{asy:power_control}
    \begin{aligned}
        p_k=&p_k^{\text{max}}\cdot\beta_k\cdot\frac{\Omega}{s^k+\Omega}\\
        &+p_k^{\text{max}}\cdot(1-\beta_k)\cdot\frac{\Theta(\Delta\boldsymbol{w}_k^t,\boldsymbol{w}_g^t-\boldsymbol{w}_g^{t-1})+1}{2}\\
        =&p_k^{\text{max}}(\beta_k\cdot \rho_k+(1-\beta_k)\cdot{\theta}_k),
    \end{aligned}
\end{equation}
where $s_k$ is the staleness factor of local model $\boldsymbol{w}_k$ at each round, $\theta_k$ is the interference factor of local model, $\Omega$ is a constant to limit the maximum degree of latency, and $\Theta(\boldsymbol{a},\boldsymbol{b})\in[-1, 1]$ represents the cosine of the angle between two vector $\boldsymbol{a}$ and $\boldsymbol{b}$.
$\beta_k\in[0,1]$ is a hyper-parameter that can make a trade-off between the staleness factor $\rho_k$ and the interference factor $\theta_k$ \cite{asynchronous}, and makes $p_k$ still subject to the individual transmit power condition \eqref{Eq:individual_power} where $0\le p_k\le p_k^{\text{max}}$.
% Based on \eqref{asy:power_control}, we can reformulate the \textbf{P1} to follow form:

By substituting the expression for $p_k$ into the original optimization problem \textbf{P1} and representing it in matrix form, we obtain the final optimization problem:
% \begin{subequations}\label{opt2:problem}
%     \begin{align}
%     &\begin{aligned}\label{opt2:objective}
%     \mathbf{P2}:\ \ {\min_{\alpha}}&\ \ 
%     \frac{L\varepsilon ^2K\sum\nolimits_{k=1}^K{{b_k^2(p_k^{\text{max}})^2(\alpha s_k+(1-\alpha)\theta_k)}^2}}{\left( \sum\nolimits_{k\in \mathcal K}{b_kp_k^{\text{max}}(\alpha s_k+(1-\alpha)\theta_k)} \right) ^2}\\
%     &+\frac{2Ld\sigma _{u}^{2}}{\left( \sum\nolimits_{k\in \mathcal K}{b_kp_k^{\text{max}}(\alpha s_k+(1-\alpha)\theta_k)} \right) ^2}
%     \end{aligned}\\
%     &\ \ \ \ \ \ \ \ \ \ {\text{s.t.}}\ \  b_k \in \left\{ 0,1 \right\}, \label{opt2:a}\\
%     &\ \ \ \ \ \ \ \ \ \ \ \ \ \ \ \ \alpha\in[0,1],\label{opt2:alpha}
%     \end{align}
% \end{subequations}
\begin{subequations}\label{opt2:problem}
    \begin{align}
    \mathbf{P2}:\ \min_{\boldsymbol{\beta}}\ &\frac{\left( \boldsymbol{\theta }+\boldsymbol{D\beta } \right) ^T\boldsymbol{P}_{\max}^{T}\mathbf{\Theta }\boldsymbol{P}_{\max}\left( \boldsymbol{\theta }+\boldsymbol{D\beta } \right) +2Ld\sigma _{u}^{2}}{\left( \boldsymbol{\theta }+\boldsymbol{D\beta } \right) ^T\boldsymbol{P}_{\max}^{T}\boldsymbol{II}^T\boldsymbol{P}_{\max}\left( \boldsymbol{\theta }+\boldsymbol{D\beta } \right)}\nonumber\\ =&\frac{\boldsymbol{\beta}^T\boldsymbol{G}\boldsymbol{\beta}+\boldsymbol{g}^T\boldsymbol{\beta}+g_0}{\boldsymbol{\beta}^T\boldsymbol{Q}\boldsymbol{\beta}+\boldsymbol{q}^T\boldsymbol{\beta}+q_0}=\frac{h_1(\boldsymbol {\beta})}{h_2(\boldsymbol{\beta})}\label{opt2:objective}\\
    {\text{s.t.}}\ \  &\beta_k\in[0,1],k=1, \cdots, K,\label{opt2:beta}
    \end{align}
\end{subequations}
where $\boldsymbol {\rho}^T=[\rho_1,\cdots,\rho_K]$, $\boldsymbol{\theta}^T = [\theta_1,\cdots,\theta_K]$, $\boldsymbol \beta^T = [\beta_1,\cdots,\beta_K]$, $\boldsymbol P^{\max}=\mathrm{diag}\{p_1^{\max}, \cdots, p_K^{\max} \}$ and $\boldsymbol D=\mathrm{diag}\{\rho_1-\theta_1, \cdots, \rho_K-\theta_K\}$. $\boldsymbol{Q}\hspace{-0.15cm}=\hspace{-0.15cm}\boldsymbol{D}^T\boldsymbol{P}_{\max}^T\boldsymbol{bb}^T\boldsymbol{P}_{\max}\boldsymbol{D}$ is a $K\times K$ symmetric positive definite matrix, $\boldsymbol{G}\hspace{-0.15cm}=\hspace{-0.15cm}\boldsymbol{D}^T\boldsymbol{P}_{\max}^T\boldsymbol{\Theta}^T\boldsymbol{P}_{\max}\boldsymbol{D}$ is a $K\times K$ symmetric positive semi-definite definite matrix, $\boldsymbol{q}\hspace{-0.15cm}=\hspace{-0.15cm}2\boldsymbol{\theta}^T\boldsymbol{P}_{\max}^T\boldsymbol{bb}^T\boldsymbol{P}_{\max}\boldsymbol{D}$, $\boldsymbol{g}\hspace{-0.15cm}=\hspace{-0.15cm}2\boldsymbol{\theta}^T\boldsymbol{P}_{\max}^T\boldsymbol{\Theta}^T\boldsymbol{P}_{\max}\boldsymbol{D}$ are $K$-vectors.
$q_0 \hspace{-0.15cm}=\hspace{-0.15cm}\boldsymbol{\theta}^T\boldsymbol{P}_{\max}^T\boldsymbol{bb}^T\boldsymbol{P}_{\max}\boldsymbol{\theta}$, $g_0 \hspace{-0.15cm}=\hspace{-0.15cm}\boldsymbol{\theta}^T\boldsymbol{P}_{\max}^T\boldsymbol{\Theta}^T\boldsymbol{P}_{\max}\boldsymbol{\theta}+2Ld\sigma_u^2$ are constants.

Problem \eqref{opt2:problem} is a nonlinear fractional programming problem, where both the dividend and divisor are convex quadratic functions subject to linear constraints. To solve this problem, we adopt an improved version of the Dinkelbach's algorithm \cite{Dinkel}, as shown in \textbf{Algorithm 2}.
The Dinkelbach's transform of problem \textbf{P2} can be formulated as:
\begin{subequations}\label{opt3:problem}
    \begin{align}
    \mathbf{P3}:\ \ \max_{\boldsymbol{\beta}}\ \ &F(\boldsymbol{\beta};\lambda)={h_2(\boldsymbol{\beta})}-\lambda {h_1(\boldsymbol {\beta})}\label{opt2:objective}\\
    {\text{s.t.}}\ \  &\beta_k\in[0,1],k=1, \cdots, K.\label{opt2:beta}
    \end{align}
\end{subequations}
where $\lambda >0$ is treated as a parameter. 
% The difficulty associated with the Algorithm 2 is
% It can be seen that 
And \textbf{P3} is a maximization of a non-concave quadratic function. 
% By introducing nonsingular matrix $\boldsymbol D_1$ and orthogonal matrix $\boldsymbol D_2$ respectively satisfying $\boldsymbol G = \boldsymbol{D}_1^T\boldsymbol{D}_1$ and $\boldsymbol{D}_2^T{\boldsymbol{D}_1^{-1}}^T\boldsymbol{Q}{\boldsymbol{D}_1^{-1}}\boldsymbol{D}_2=\boldsymbol{\Omega}$, where $\boldsymbol\Omega=\text{diag}\{\omega_i\}$ and $\boldsymbol{D}_2^T\boldsymbol{D}_2=\boldsymbol{E}$, the objective of \textbf{P3} is equivalent to
% \begin{equation}
% \begin{aligned}
%     F(\boldsymbol{\beta};\lambda) =& \boldsymbol{\beta}^T\boldsymbol{D}_2^T\boldsymbol{D}_1^T(\boldsymbol{\Omega}-\lambda \boldsymbol{E})\boldsymbol{D}_1\boldsymbol{D}_2\boldsymbol{\beta}\\
%     &+(\boldsymbol{q}^T-\lambda \boldsymbol{g}^T)\boldsymbol{\beta}+(q_0-\lambda g_0)
% \end{aligned}
% \end{equation}
Now, we apply the standard piecewise linear approximation of quadratic functions and reformulate \textbf{P3} as a 0-1 linear integer programming problem. Since $\boldsymbol{G}$ is symmetric and positive definite, there exists a nonsingular matrix $\boldsymbol{M}_1$ such that $\boldsymbol{G}=\boldsymbol{M}_1^T\boldsymbol{M_1}$. Let $\boldsymbol{\beta}=\boldsymbol{M}_1\boldsymbol{\beta}$. Then
\begin{equation}
    F(\boldsymbol{\beta};\lambda)=\boldsymbol{y}^T \boldsymbol{S}\boldsymbol{y}-\lambda\boldsymbol{y}^T\boldsymbol{y}+(\boldsymbol{q}^T-\lambda\boldsymbol{g}^T)\boldsymbol{\beta}+(q_0-\lambda g_0),
\end{equation}
and since $\boldsymbol{Q}$ is symmetric and positive semi-definite, there exists an orthogonal matrix $\boldsymbol{M}_2$ such that $\boldsymbol{M}_2^T\boldsymbol{S}\boldsymbol{M}_2=\boldsymbol{N}$, $\boldsymbol{M}_2^T\boldsymbol{M}=\boldsymbol{I}$, where $\boldsymbol{N}=\text{diag}(n_i)$. Therefore, we have
\begin{equation}
    \begin{aligned}
    F(\boldsymbol{\beta};\lambda)=&\boldsymbol{\beta}^T\boldsymbol{M}_1^T\boldsymbol{M}_2 (\boldsymbol{N}-\lambda\boldsymbol{I})\boldsymbol{M}_2^T\boldsymbol{M}_1\boldsymbol{\beta}\\
    &+(\boldsymbol{q}^T-\lambda\boldsymbol{g}^T)\boldsymbol{\beta}+(q_0-\lambda g_0),
    \end{aligned}
\end{equation}

Problem \textbf{P3} can be represented as follow:
\begin{subequations}\label{opt4:problem}
    \begin{align}
    \mathbf{P4}:\ \ \max_{\boldsymbol{\beta}}\ \ &f(\boldsymbol{\beta};\lambda)=\boldsymbol{z}^T(\boldsymbol{N}-\lambda\boldsymbol{I})\boldsymbol{z}\notag\\
    &+(\boldsymbol{q}^T-\lambda\boldsymbol{g}^T)\boldsymbol{M}^{-1}\boldsymbol{z}+(q_0-\lambda g_0)\\
    {\text{s.t.}}\ \  &\beta_k\in[0,1],k=1, \cdots, K,\label{opt4:beta}\\
    &\boldsymbol{z}=\boldsymbol{M}\boldsymbol{\beta}\label{opt4:transform},
    \end{align}
\end{subequations}
where $\boldsymbol{M}=\boldsymbol{M}_2^T\boldsymbol{M}_1$. Now, let 
\begin{equation}
    \boldsymbol{Z}=\{\boldsymbol{z}\in\mathbb{R}^n|\boldsymbol{M}^{-1}\boldsymbol{z}=\boldsymbol{\beta}
    \}
\end{equation}
and let
\begin{align}
    &z_{i\varrho +1}=\max\{{z}_i|\boldsymbol{z}\in\boldsymbol{Z}\},i = 1,2,\cdots,n,\\
    &z_{1}=\min\{{z}_i|\boldsymbol{z}\in\boldsymbol{Z}\},i = 1,2,\cdots,n,
\end{align}

We split the interval $[z_{i1},z_{i\varrho+1}]$ into m sub-intervals of equal length. The piecewise linear approximation of the objective function can be represented by introducing a number of auxiliary variables $\gamma_{ij}$, $i=1,2,\cdots,n$ and $j=1,2,\cdots, \varrho+1$, as follows:
\begin{equation}\label{quadratic}
    \begin{aligned}
        \boldsymbol{z}^T(\boldsymbol{N}-\lambda\boldsymbol{I})\boldsymbol{z}=&-\sum_{i=1}^{n-h} c_i(\lambda)\sum_{j=1}^{\varrho+1} z_{ij}^2\gamma_{ij} \\
        &+\sum_{i=n-h+1}^{n} c_i(\lambda)\sum_{j=1}^{\varrho+1} z_{ij}^2 \gamma_{ij}
    \end{aligned}
\end{equation}
\begin{align}
    &z_i=\sum_{j=1}^{\varrho+1}z_{ij}\gamma_{ij}, i=1,2,\cdots,n,\label{pla1}\\
    &\sum_{j=1}^{\varrho+1}\gamma_{ij}=1, i=1,2,\cdots,n,\label{pla2}\\
    &\gamma_{ij}\geq 0,i=1,2,\cdots,n and j=1,2,\cdots, \varrho+1 \label{pla3}
\end{align}

Let us note that we need to introduce the 0-1 variables $c_{ij}$, $i=n-j+1, \cdots, n$ for $j=1,2,\cdots,\varrho$,
\begin{equation}\label{01var}
    \begin{aligned}
        &\gamma_{i1}\leq c_{i1},\\
        &\gamma_{i2}\leq c_{i1}+c_{i2},\\
        &\vdots\\
        &\gamma_{i\varrho}\leq c_{ih-1}+c_{ih},\\
        &\sum_{j=1}^m=y_{ij}=1,
    \end{aligned}
\end{equation}

Therefore, the 0-1 linear integer programming problem can be reformulated as follow:
\begin{subequations}\label{opt5:problem}
    \begin{align}
    \max_{\boldsymbol{z}}\ \ &-\sum_{i=1}^{n-h} c_i(\lambda)\sum_{j=1}^{\varrho+1} z_{ij}^2\gamma_{ij} \notag
    +\sum_{i=n-h+1}^{n} c_i(\lambda)\sum_{j=1}^{\varrho+1} z_{ij}^2 \gamma_{ij}\notag\\
    &+(\boldsymbol{q}^T-\lambda\boldsymbol{g}^T)\boldsymbol{M}^{-1}\boldsymbol{z}+(q_0-\lambda g_0)\\
    {\text{s.t.}}\ \  &\boldsymbol{z}\in\boldsymbol{Z},\\
    &\eqref{quadratic},\eqref{pla1},\eqref{pla2},\eqref{pla3},\eqref{01var}.\notag
    \end{align}
\end{subequations}

This is a mixed integer programming problem with $h \times \varrho$ 0-1 variables which can be solved by IBM CPLEX Optimizer efficiently. And then we can get the optimal $\boldsymbol{\beta}$ by using \eqref{opt4:transform} and the solution of the problem \textbf{P1} by adopting \eqref{asy:power_control}.

% \begin{subequations}\label{opt3:problem}
%     \begin{align}
%     \mathbf{P3}:\ \ \max_{\boldsymbol{\beta}}\ \ &\frac{\boldsymbol{\beta}^T\boldsymbol{Q}\boldsymbol{\beta}+\boldsymbol{q}^T\boldsymbol{\beta}+q_0}{\boldsymbol{\beta}^T\boldsymbol{G}\boldsymbol{\beta}+\boldsymbol{g}^T\boldsymbol{\beta}+g_0}\label{opt2:objective}\\
%     {\text{s.t.}}\ \  &\beta_k\in[0,1],k=1, \cdots, K,\label{opt2:beta}
%     \end{align}
% \end{subequations}
% Associated with the problem \textbf{P3}, we define a function 
% \begin{equation}
% \begin{aligned}
% F(\boldsymbol{\beta};\lambda)=& {\boldsymbol{\beta}^T\boldsymbol{Q}\boldsymbol{\beta}+\boldsymbol{q}^T\boldsymbol{\beta}+q_0}\\
% &-\lambda({\boldsymbol{\beta}^T\boldsymbol{G}\boldsymbol{\beta}+\boldsymbol{g}^T\boldsymbol{\beta}+g_0})
% \end{aligned}
% \end{equation}
% where $\lambda>0$ is treated as a parameter. The 

% As part of the solution process, a nonlinear quadratic programming problem arises, which we convert into a mixed integer optimization problem using piecewise linear approximation. We then use an optimization solver to solve the resulting problem.

% Thus, we transform the original problem \textbf{P1} into an optimization problem with a single parameter. We obtain the optimal solution $\alpha^*$ of \textbf{P2} by numerical traversal method, and then minimize the value of the objective function. The whole process of PAOTA is shown in \textbf{Algorithm 1}.

\begin{algorithm}[t]
\caption{Dinkelbach's Method}
\label{alg:Dinkelbach}
\begin{algorithmic}[1] %这个1 表示每一行都显示数字
\REQUIRE %算法的输入参数：Input
    tolerance $\varepsilon$; $\lambda_0$ satisfying $F(\boldsymbol{\beta};\lambda_0)\geq 0$; iteration number $n=0$;\\
    \REPEAT 
    \STATE set $\lambda=\lambda_n$
    \STATE Solve 0-1 MIP Problem \eqref{opt5:problem} derived from \eqref{opt3:problem}'s piecewise linear approximation to obtain $\boldsymbol{\beta}^*$;
    \STATE $\lambda_{n+1}\leftarrow \tfrac{h_2(\boldsymbol{\beta}^*)}{h_1(\boldsymbol{\beta}^*)}$;
    \STATE $n\leftarrow n+1$;
    \UNTIL{$F(\boldsymbol{\beta};\lambda_n)\le \varepsilon$}
    % \vspace{-0.3cm}
\end{algorithmic}
\end{algorithm}
% \clearpage
\section{Simulation Results}
% \begin{table}[b]
% \vspace{-0.3cm}
% \renewcommand\arraystretch{1.25}
% \caption{Parameters Settings}
% \begin{center}
% \begin{tabular}{c|c|c}
% \hline
% \textbf{Symbol}& \textbf{Description}& \textbf{Value} \\
% \hline
% $K$& Number of clients  & 100\\
% \hline
% $R$& Number of global rounds & 200\\
% \hline
% $E$& Number of local iteration& 5\\
% \hline
% $\sigma_u$& Uplink noise variance & 0.1\\
% \hline
% $h_k^r$ & Channel gain of uplink transmission & $\mathcal{CN}(0,1)$\\
% \hline
% $\Delta T$& Aggregation peroid of PAOTA & 0.8 ms\\
% \hline
% \end{tabular}
% \label{tab1}
% \end{center}
% \end{table}
\subsection{Experiment Settings}
% In this section, we evaluate the performances of the presented algorithm PAOTA for the task of image classification.
% We train a multi-layer perception (MLP) network which has two hidden layers with 10 hidden nodes, and run the task on the MNIST dataset \cite{mnist}. 
% We consider the data distribution across the clients is non-IID distribution, where the number of training samples in different clients varies from $\{300, 600, 900, 1200, 1500\}$ and each client contains 5 categories of digit images at most. The parameters related to the wireless FL system are shown in TABLE \uppercase\expandafter{\romannumeral1}.
We consider a cellular network consisting of a basic station (BS) and 100 clients participating in FEEL training, where the downlink transmission is error-free. The maximal transmit power of local devices is $p_{\max}^k=15 $w.  We set the uplink transmission bandwidth as 20MHz, and the channel noise power spectral density as $N_0=-174$dBm/Hz. And we set $M = 5$, $L = 10$, and $\Omega= 3$.
% The other parameters used in simulation are listed in TABLE \uppercase\expandafter{\romannumeral1}. 
We train a multi-layer perception (MLP) network which has two hidden layers with 10 nodes on the MNIST dataset. Considering the non-IID distribution among each clients, we set the number of training samples in different clients varies from $\{300, 600, 900, 1200, 1500\}$ and each device contains five categories of digit images at most. 

% In practice, clients in the FL system may need to perform other higher priority tasks, and provide less computation resources in FL local training tasks.
In order to realize the heterogeneity of edge devices in the computing ability, we set the computation latency of each client during different local training round to follow the uniform distribution $\mathcal U(5, 15)$s, and the period of model aggregation at each epoch for PAOTA as $\Delta T=8$s.

\subsection{Performance Comparison}
% \begin{figure}[t]
% \centerline{\includegraphics[width=0.5\textwidth]{draft/figures/test_acc.pdf}}
% \caption{Test accuracy in non-IID settings.}
% \label{fig:test_acc}
% \vspace{-0.3cm}
% \end{figure}
% \begin{figure}[t]
% \centerline{\includegraphics[width=0.5\textwidth]{draft/figures/train_loss.pdf}}
% \caption{Train loss in non-IID settings.}
% \label{fig:train_loss}
% \vspace{-0.3cm}
% \end{figure}

We compare the proformance of PAOTA algorithm with the following federated learning algorithms:

(1) Local SGD \cite{b1}: An ideal synchronous federated learning algorithm, where each user transmits its local model without considering transmission loss.

(2) COTAF \cite{T2020OTA}: One of the classic AirComp based FEEL algorithms where each user transmits its model updates through the MAC and performs time-varying pre-coding.

For fairness consideration, we set an equal number of participating clients for each round of training in the three algorithms.
To verify the effectiveness of the proposed algorithm, we first numerically evaluate the gap between expected objective and optimal loss function value, i.e., $\mathbb{E}[F(\boldsymbol{w}^r)]-F(\boldsymbol{w}^*)$. 

\begin{figure}[t]
    \vspace{-0.3cm}
    \centerline{\includegraphics[width=0.5\textwidth]{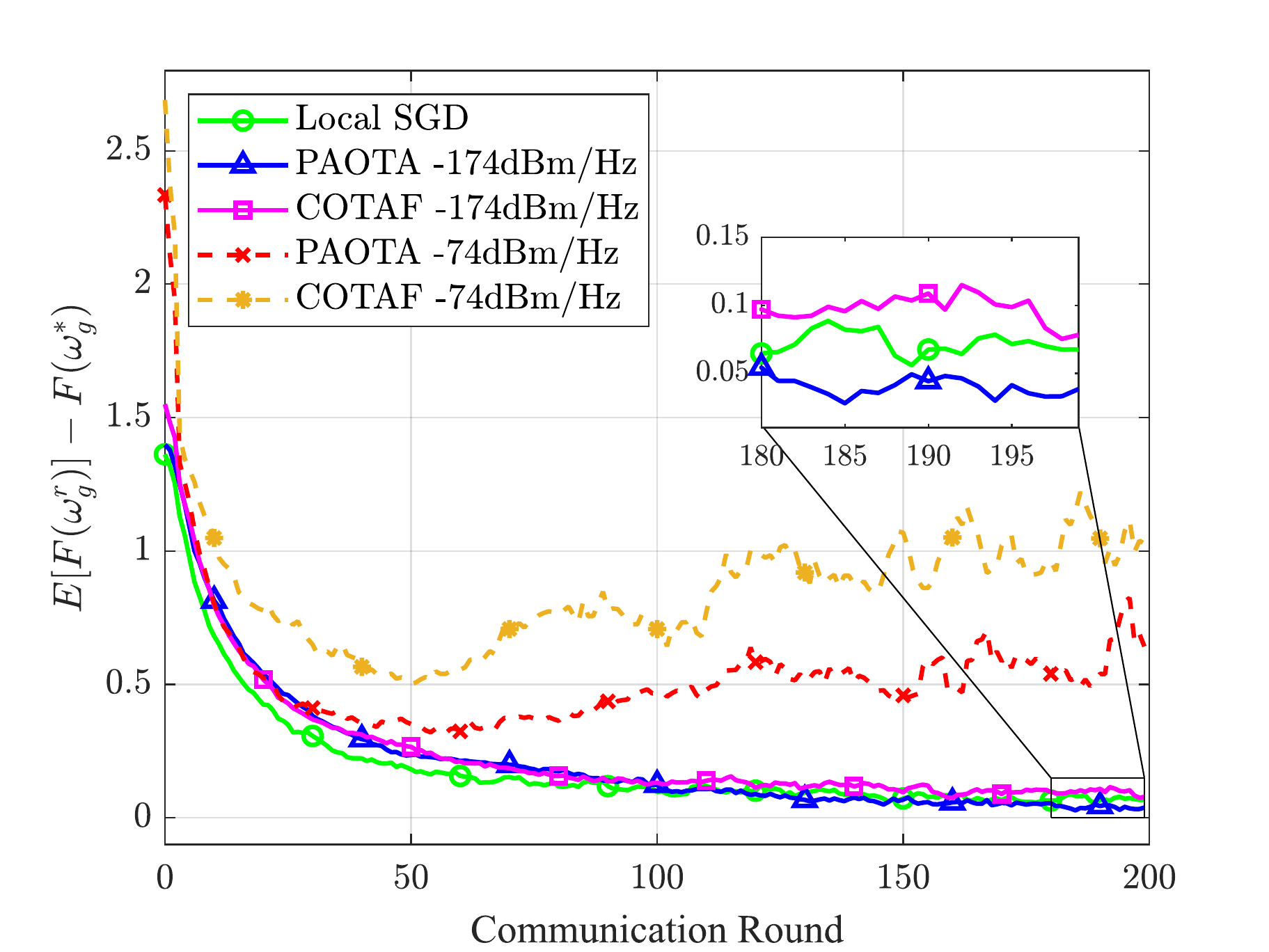}}
    \caption{Train loss in non-IID settings with bandwidth $B$=20MHz.}
    \label{fig:convergence}
    % \vspace{-0.5cm}
\end{figure}

As shown in Fig. \ref{fig:convergence}, we observe that PAOTA can achieve convergence speed close to Local SGD when $N_0 = -174d$Bm/Hz. This verifies that the PAOTA algorithm can effectively compensate for the negative impact of additive noise and asynchronous aggregation mechanism on model convergence speed. Moreover, as the number of iterations increases, PAOTA can achieve a smaller gap than Local SGD, indicating that PAOTA improves the utilization of heterogeneous data in clients through the proposed semi-asynchronous aggregation strategy.

Furthermore, in Fig. \ref{fig:convergence}, both COTAF and PAOTA can achieve similar convergence performance when $N_0 = -174$dBm/Hz. 
However, when the noise power spectral density is increased to -74dBm/Hz while ensuring that the uplink transmission power of clients remains constant, PAOTA is more robust than COTAF. This is because when optimizing transmission power, PAOTA considers the additive noise parameters of the channel in optimization and thus implements uplink power control adaptively.

\begin{figure}[t]
\vspace{-0.3cm}
\centerline{\includegraphics[width=0.55\textwidth]{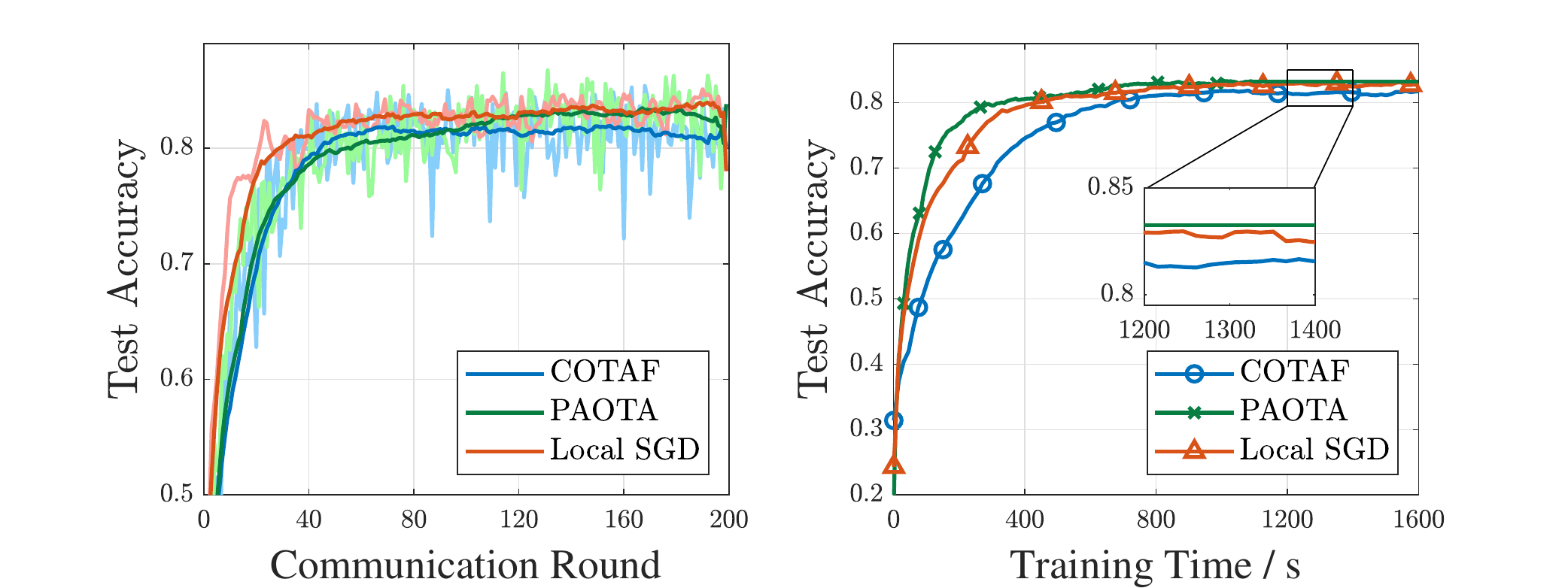}}
\caption{Test accuracy in non-IID settings.}
\label{fig:acc}
\vspace{-0.3cm}
\end{figure}

Fig. \ref{fig:acc} compares the test set accuracy of the three algorithms with respect to communication rounds and training time where we set $N_0=-174$dBm/Hz. It can be observed that PAOTA ultimately achieved a prediction accuracy of 83.5\%, which is 1.1\% higher than that of Local SGD; the prediction accuracy under COTAF is 81\%, which is lower than the ideal situation due to the negative impact of the wireless channel on the model accuracy. 

From the perspective of training time, the global iteration training time of PAOTA is set to $\Delta T$, while the global iteration training time of Local SGD and COTAF is determined by the client with the longest local computing time in this global round. 
We list the time and rounds required to achieve the target accuracy algorithms in Table \uppercase\expandafter{\romannumeral1}. 
It can be seen that PAOTA requires more rounds than Local SGD to achieve the target accuracy. However, since the time per round is fixed for PAOTA, while Local SGD needs to wait for all selected clients to complete training before aggregation, PAOTA spends less time to achieve the same target accuracy. For example, PAOTA saves 25\% of time to achieve the target accuracy 80\% than Local SGD. PAOTA achieves the performance improvement mentioned above because its semi-asynchronous aggregation mechanism can avoid the generation of bottleneck nodes for each global round. Additionally, it controls the weights of expired local models in aggregation to ensure the utilization of data in clients with long local training time.
% shows the number of training rounds and the training time taken by different algorithms to achieve different target test dataset accuracy. 
% Compared with TT-Fed, it takes slightly more global training rounds for PAOTA to achieve the same accuracy value. However, because the aggregation cycle time under PAOTA is shorter than TT-Fed, the time spent for PAOTA is less.
% Since FedAvg selects the same number of clients in each round randomly, it will lead to a long time for one communication round once there are stragglers in selected clients.
% FedAsync can learn quickly because of the frequent global aggregation, but it cannot attain a ideal performance in highly non-IID settings.
\addtolength{\topmargin}{0.05in}

% \clearpage
\section{Conclusion}
In this paper, we first propose a semi-asynchronous mechanism called PAOTA under wireless MAC channels, where the clients suffer from computing heterogeneity. 
Then, we analyze the convergence behavior of PAOTA and illustrate how the asynchronous strategy and the wireless transmission affect the upper bound of the expected gap between the expected and optimal global loss. 
Considering the staleness discount introduced by the asynchronous mechanism and the interference effect of local client’s update, we model this trade-off problem as the transmission power optimization. 
In non-IID settings, the simulation results demonstrate that PAOTA can achieve a better performance than other benchmarks in terms of the robustness in terrible wireless condition and convergent speed.

\begin{table}[t]
    \renewcommand\arraystretch{1.25}
    \caption{Convergence Time}
    \begin{center}
    \begin{tabular}{c|c|c|c|c|c}
    \hline
    \multicolumn{2}{c|}{\textbf{Target Accuracy}}& \textbf{50\%} & \textbf{60\%} & \textbf{70\%} &\textbf{80\%}\\
    \hline
    \multirow{2}{*}{\textbf{PAOTA}} & round & 6 & 10 & 18 & 57\\
    \cline{2-6}
     & time/s  & \textbf{36} & \textbf{60} & \textbf{108} & \textbf{342}\\
    \hline
    \multirow{2}{*}{\textbf{Local SGD}} & round & 3 & 6 & 12 & 29\\
    \cline{2-6}
     & time/s  & 45.61 & 78.17 & 181.24 & 451.62\\
     \hline
    \multirow{2}{*}{\textbf{COTAF}} & round & 6 & 12 & 29 & 44\\
    \cline{2-6}
     & time/s  & 91.30 & 181.21 & 316.75 & 676.93\\
    \hline
    % \hline
    % \multirow{2}{*}{MultiRow}&\multicolumn{2}{|c|}{\textbf{PAOTA}} &\multicolumn{2}{|c|}{\textbf{PAOTA}} &\multicolumn{2}{|c|}{\textbf{PAOTA}} &\multicolumn{2}{|c|}{\textbf{PAOTA}} \\
    % \cline{2-9} 
    %  & round & time & round & time & round & time & round & time  \\
    % \hline
    % copy& 3 &  &  &  &  &  &  &    \\
    % \multicolumn{4}{l}{$^{\mathrm{a}}$Sample of a Table footnote.}
    \end{tabular}
    \label{tab2}
    \end{center}
    \vspace{-0.5cm}
\end{table}

\clearpage
{\appendices
\section{proof of lemma1}
Firstly, we derive the upper bound of the expected square norm of the difference between the local updated model at each SGD iteration and the previous global model $\mathbb{E} [ \| \boldsymbol{w}_g^{ r-s_k^r } -\boldsymbol{w}_{k,{\tau -1}}^{ r-s_k^r} \| _{2}^{2} ]$ as follow:
\begin{align}\label{prooflemma1-1}
    &\mathbb{E} [ \| \boldsymbol{w}_g^{r-s_k^r}  -\boldsymbol{w}_{k,\tau -1}^{{r-s_k^r}}      \| _{2}^{2} ]\notag\\
    =&\mathbb{E}   [   \| \eta _t\sum\nolimits_{l=1}^{\tau -1}{\nabla F_k  ( \boldsymbol{w}_{k,l-1}^{{r-s_k^r}}      ;D_{k,l}^{{r-s_k^r}}      )}  \| _{2}^{2}  ]\notag\\
    \le & {\eta _t}^2  ( \tau -1  )\sum\nolimits_{l=1}^{\tau -1}{\mathbb{E}   [   \| \nabla F_k  ( \boldsymbol{w}_{k,l-1}^{{r-s_k^r}};D_{k,l}^{{r-s_k^r}})  \| _{2}^{2}  ]}\notag\\
    \mathop {\mathrm{=}}^{(a)}&{\eta _t}^2  ( \tau -1  )\sum\nolimits_{l=1}^{\tau -1}\mathbb{E}  [ \| \nabla F_k  ( \boldsymbol{w}-_{k,l-1}^{{r-s_k^r}};D_{k,l}^{{r-s_k^r}})\notag\\
    &-\nabla F_k  ( \boldsymbol{w}_{k,l-1}^{{r-s_k^r}} ) \| _{2}^{2}  ]\notag\\
    &+{\eta _t}^2  ( \tau -1  ) \sum\nolimits_{l=1}^{\tau -1}{\mathbb{E}   [   \| \nabla F_k  ( \boldsymbol{w}_{k,l-1}^{{r-s_k^r}}       )  \| _{2}^{2}  ]}\notag\\
    \mathop {=}^{(b)}&{\eta _t}^2  ( \tau -1  ) ^2\sigma ^2\notag\\
    &+{\eta _t}^2  ( \tau -1  ) \sum\nolimits_{l=1}^{\tau -1}{\mathbb{E}   [   \| \nabla F_k  ( \boldsymbol{w}_{k,l-1}^{{r-s_k^r}}       )  \| _{2}^{2}  ]}\notag\\
    % \le& {\eta _t}^2M^2\sigma ^2+{\eta _t}^2M\sum\nolimits_{l=1}^{\tau -1}{\mathbb{E}   [   \| \nabla F_k  ( \boldsymbol{w}_{k,l-1}^{{r-s_k^r}}      )  \| _{2}^{2}  ]}\notag\\
    \le&{\eta _t}^2M^2\sigma ^2+{\eta _t}^2M\sum\nolimits_{l=1}^{\tau -1}\mathbb{E}  [ \| \nabla F_k  ( \boldsymbol{w}_{k,l-1}^{{r-s_k^r}}       )\notag \\
    &-\nabla F_k  ( \boldsymbol{w}_g^{r-s_k^r}      ) +\nabla F_k  ( \boldsymbol{w}_g^{r-s_k^r}   ) \| _{2}^{2}  ]\notag\\
    \mathop {\le}^{(c)}&{\eta_t}^2M^2\sigma ^2+2{\eta _t}^2M\sum\nolimits_{l=1}^{\tau -1}{\mathbb{E}   [   \| \nabla F_k  ( \boldsymbol{w}_g^{r-s_k^r}   )  \| _{2}^{2}  ]}\notag\\
    &+2{\eta _t}^2M\sum\nolimits_{l=1}^{\tau -1}\mathbb{E}  [ \| \nabla F_k  ( \boldsymbol{w}_{k,l-1}^{{r-s_k^r}}       ) -\nabla F_k  ( \boldsymbol{w}_g^{r-s_k^r}   ) \| _{2}^{2}  ]\notag\\
    % &\notag\\
    \mathop {\le}^{(d)}&{\eta _t}^2M^2\sigma ^2\notag\\
    &+2{\eta _t}^2L^2M\sum\nolimits_{l=1}^{\tau -1}{\mathbb{E}   [   \| \boldsymbol{w}_{k,l-1}^{{r-s_k^r}}      -\boldsymbol{w}_g^{r-s_k^r}   \| _{2}^{2}  ]}\notag\\
    &+2{\eta _t}^2M\sum\nolimits_{l=1}^{\tau -1}\mathbb{E}  [ \| \nabla F_k  ( \boldsymbol{w}_g^{r-s_k^r}   ) \notag\\
    &-\nabla F_g  ( \boldsymbol{w}_g^{r-s_k^r}   ) +\nabla F_g  ( \boldsymbol{w}_g^{r-s_k^r}   ) \| _{2}^{2}  ]\notag\\
    \mathop {\le}^{(e)}&{\eta _t}^2M^2\sigma ^2+2{\eta _t}^2L^2M\sum\nolimits_{l=1}^{\tau -1}{\mathbb{E}   [   \| \boldsymbol{w}_{k,l-1}^{{r-s_k^r}}  -\boldsymbol{w}_g^{r-s_k^r}   \| _{2}^{2}  ]}\notag\\
    % &\notag\\
    &+4{\eta _t}^2M\sum\nolimits_{l=1}^{\tau -1}\mathbb{E}  [ \| \nabla F_k  ( \boldsymbol{w}_g^{r-s_k^r}   ) -\nabla F_g  ( \boldsymbol{w}_g^{r-s_k^r}   ) \| _{2}^{2}\notag\\
    &+  \| \nabla F_g  ( \boldsymbol{w}_g^{r-s_k^r}   )  \| _{2}^{2}  ]\notag\\
    % \mathop {\le}^{(f)}&{\eta _t}^2M^2\sigma ^2+4{\eta _t}^2M^2\zeta +4{\eta _t}^2M^2\mathbb{E}   [   \| \nabla F_g  ( \boldsymbol{w}_g^{r-s_k^r}   )  \| _{2}^{2}  ]\notag\\
    % &+2{\eta _t}^2L^2M\sum\nolimits_{l=1}^{\tau -1}{\mathbb{E}   [   \| \boldsymbol{w}_{k,l-1}^{{r-s_k^r}}  -\boldsymbol{w}_g^{r-s_k^r}   \| _{2}^{2}  ]}\notag\\
    \mathop {\le}^{(f)}&{\eta _t}^2M^2\sigma ^2+4{\eta _t}^2M^2\zeta +4{\eta _t}^2M^2\beta ^2\mathbb{E}   [   \| \nabla F_g  ( \boldsymbol{w} _g^r  )  \| _{2}^{2}  ]\notag\\
    &+2{\eta _t}^2L^2M\sum\nolimits_{l=1}^{\tau -1}{\mathbb{E}   [   \| \boldsymbol{w}_{k,l-1}^{{r-s_k^r}}  -\boldsymbol{w}_g^{r-s_k^r}   \| _{2}^{2}  ]}
\end{align}
where equality (a) is due to 
\begin{equation}\label{square expectation decomposition}
    \mathbb{E} [   \| \boldsymbol{x}  \| ^2 ] =\mathbb{E}[ \| \boldsymbol{x}-\mathbb{E} [ \boldsymbol{x} ] \| ^2 ] +\| \mathbb{E} [ \boldsymbol{x} ] \| ^2 ,
\end{equation} 
and 
\begin{equation}\label{gradient expectation}
    \mathbb{E}   [ \nabla F_k  ( \boldsymbol{w}_{k,l-1}^{{r-s_k^r}}  ;D_{k,l}^{{r-s_k^r}}       )  ] =\nabla F_k  ( \boldsymbol{w}_{k,l-1}^{{r-s_k^r}}   ) ,
\end{equation}
equality (b) is due to the equation 
\begin{equation}\label{gradient variance}
\mathbb{E} [ \| \nabla F_k( \boldsymbol{w}_{k,l-1}^{{r-s_k^r}}  ;D_{k,l}^{{r-s_k^r}}) -\nabla F_k( \boldsymbol{w}_{k,{l-1}}^{r-s_k^r}) \| ^2 ] =\sigma ^2,
\end{equation}
equality (c) and equality (e) are both due to 
% \begin{equation}
%     \mathbb{E} [   \| \boldsymbol{x}_1+\boldsymbol{x}_2  \| ^2 ] =2\mathbb{E} [\| \boldsymbol{x}_1\| ^2] +2\mathbb{E} [\| \boldsymbol{x}_2  \| ^2 ], 
% \end{equation}
\begin{equation}\label{sum square}
    \| \boldsymbol{x}_1+\boldsymbol{x}_2 \| _{2}^{2}\le 2\| \boldsymbol{x}_1 \| _{2}^{2}+2\| \boldsymbol{x}_2 \| _{2}^{2}
\end{equation}
equality (d) is by the inequality \eqref{ass1-2} in Assumption 1, equality (f) is by the inequality \eqref{ass2-1} in Assumption 2 and the \eqref{ass3-3} in Assumption 3.

Then, summing both sides of \eqref{prooflemma1-1} from $\tau=1$ to $M$ yields
\begin{align}\label{prooflemma1-2}
    &\sum\nolimits_{\tau =1}^M{\mathbb{E}   [   \| \boldsymbol{w}_g^{r-s_k^r}  -\boldsymbol{w}_{k,\tau -1}^{{r-s_k^r}} \| _{2}^{2}  ]}\notag\\
    \le& {\eta _t}^2M^3\sigma ^2+4{\eta _t}^2M^3\zeta +4{\eta _t}^2M^3\beta ^2\mathbb{E}   [   \| \nabla F_g  ( \boldsymbol{w} _g^r  )  \| _{2}^{2}  ]\notag\\
    &+2{\eta _t}^2L^2M\sum\nolimits_{\tau =1}^M\sum\nolimits_{l=1}^{\tau -1}\mathbb{E}  [ \| \boldsymbol{w}_{k,l-1}^{{r-s_k^r}} -\boldsymbol{w}_g^{r-s_k^r}  \| _{2}^{2}  ]\notag\\
    \le& {\eta _t}^2M^3\sigma ^2+4{\eta _t}^2M^3\zeta +4{\eta _t}^2M^3\beta ^2\mathbb{E}   [   \| \nabla F_g  ( \boldsymbol{w} _g^r  )  \| _{2}^{2}  ]\notag\\
    &+2{\eta _t}^2L^2M^2\sum\nolimits_{\tau =1}^M{\mathbb{E}   [   \| \boldsymbol{w}_{k,\tau -1}^{{r-s_k^r}}      -\boldsymbol{w}_g^{r-s_k^r}   \| _{2}^{2}  ]}\notag\\
\end{align}

Finally, rearranging the terms in \eqref{prooflemma1-2} yields Lemma 1.
\section{proof of Theorem 1}
According to \eqref{aggre-final} and the \eqref{ass1-1} in Assumption 1, we have 
\begin{equation}\label{prooftheorem1-1}
    \begin{aligned}
        F ( \boldsymbol{w}_g^{r+1}   )\le& F(\boldsymbol{w}_g^r)+< \boldsymbol{w}_g^{r+1} -\boldsymbol{w}_g^r ,\nabla F (\boldsymbol{w}_g^r)> \\
        &+\frac{L}{2} \| \boldsymbol{w}_g^{r+1}-\boldsymbol{w}_g^r \| _{2}^{2},
    \end{aligned}
\end{equation}
By taking expectation on both sides of \eqref{prooftheorem1-1}, we obtain
\begin{equation}\label{prooftheorem1-2}
    \begin{aligned}
        &\mathbb{E} [ F( \boldsymbol{w}_g^{r+1} ) ] \\
        \le& \mathbb{E} [ F( \boldsymbol{w}_g^{r} ) ] +\underset{A_1}{\underbrace{\mathbb{E} [ < \boldsymbol{w}_g^{r+1} -\boldsymbol{w}_g^{r} ,\nabla F( \boldsymbol{w}_g^{r} ) > ] }}\\
        &+\frac{L}{2}\underset{A_2}{\underbrace{\mathbb{E} [ \| \boldsymbol{w}_g^{r+1} -\boldsymbol{w}_g^{r} \| _{2}^{2} ] }}.
    \end{aligned}
\end{equation}
Then we bound the terms $A_1$ and $A_2$ in the right hand sides of \eqref{prooftheorem1-2} in the following.
\subsection{Bound of $A_1$}
In this section, we derive the bound of $A_1$,
\begin{equation}\label{prooftheorem1A1}
    \begin{aligned}
        A_1=&\mathbb{E} [ < \boldsymbol{w}_g^{r+1} -\boldsymbol{w}_g^r ,\nabla F( \boldsymbol{w}_g^r ) > ] \\
        =&\mathbb{E} [ < \tilde{\boldsymbol{w}}^r -\boldsymbol{w}_g^r +\sum\nolimits_{k=1}^K{\alpha _k^r \Delta \boldsymbol{w}_k^r}+\tilde{\boldsymbol{n}}^r ,\nabla F( \boldsymbol{w}_g^r ) > ] \\
        =&\underset{B_1}{\underbrace{\mathbb{E} [ < \tilde{\boldsymbol{w}}^r -\boldsymbol{w}_g^r ,\nabla F( \boldsymbol{w}_g^r ) > ] }}+\underset{B_3}{\underbrace{\mathbb{E} [ < \tilde{\boldsymbol{n}}^r ,\nabla F( \boldsymbol{w}_g^r ) > ] }}\\
        &+\underset{B_2}{\underbrace{\mathbb{E} [ < \sum\nolimits_{k=1}^K{\alpha _k^r \Delta \boldsymbol{w}_k^r},\nabla F( \boldsymbol{w}_g^r ) > ] }}
    \end{aligned}
\end{equation}
where $b_3=0$ because of $\tilde{\boldsymbol{n}}^r$ is orthogonal to $\nabla F( \boldsymbol{w}_g^r )$, and $\mathbb{E} [ \tilde{\boldsymbol{n}} ^r ] =0$.

We analyze $B_1$ and $B_2$ separately. We bound the term $B_1$ as ,
\begin{equation}\label{prooftheorem1B1}
    \begin{aligned}
        B_1&=\mathbb{E} [ < \sum\nolimits_{k=1}^K{\alpha _k^r ( \boldsymbol{w}_g^{r-s_k^r} -\boldsymbol{w}_g^r )},\nabla F( \boldsymbol{w}_g^r ) > ] \\
        &=\sum\nolimits_{k=1}^K{\alpha _k^r \mathbb{E} [ < \boldsymbol{w}_g^{r-s_k^r} -\boldsymbol{w}_g^r ,\nabla F( \boldsymbol{w}_g^r ) > ]}\\
        &\mathop {\le}^{(a)}\sum\nolimits_{k=1}^K{\alpha _k^r \delta \mathbb{E} [ \| \nabla F( \boldsymbol{w}_g^r ) \| _{2}^{2} ]}\\
        &=\delta \mathbb{E} [ \| \nabla F( \boldsymbol{w}_g^r ) \| _{2}^{2} ] 
    \end{aligned}
\end{equation}
where (a) is due to the \eqref{ass3-1} in Assumption 3. Then, we study the term $B_2$ as follow, 
\begin{equation}\label{prooftheorem1B2}
    \begin{aligned}
        &B_2\\
        =&\mathbb{E} [ < \sum\nolimits_{k=1}^K{\alpha _k^r \Delta \boldsymbol{w}_k^r},\\
        &\nabla F( \boldsymbol{w}_g^r ) > ] \\
        =&\mathbb{E} [ < \sum\nolimits_{k=1}^K{\alpha _k^r ( -\eta _t\sum\nolimits_{\tau =1}^M{\nabla F_k( \boldsymbol{w}_{k,\tau -1}^{{r-s_k^r}} ;D_{k,\tau}^{r-s_k^r} )} )},\\
        &\nabla F( \boldsymbol{w}_g^r ) > ] \\
        =&-\eta _t\sum\nolimits_{\tau =1}^M\mathbb{E} [ < \sum\nolimits_{k=1}^K{\alpha _k^r \nabla F_k( \boldsymbol{w}_{k,\tau -1}^{{r-s_k^r}} ;D_{k,\tau}^{r-s_k^r} )},\\
        &\nabla F( \boldsymbol{w}_g^r ) > ]\\
        % \mathop {=}^{( a )}&\tfrac{\eta _t}{2}\sum\nolimits_{\tau =1}^M\mathbb{E} [ \| \sum\nolimits_{k=1}^K{\alpha _k^r \nabla F_k( \boldsymbol{w}_{k,\tau -1}^{{r-s_k^r}} ;D_{k,\tau}^{r-s_k^r} )}\\
        % &-\nabla F( \boldsymbol{w}_g^r ) \| _{2}^{2} ]-\tfrac{\eta _t}{2}\sum\nolimits_{\tau =1}^M{\mathbb{E} [ \| \nabla F( \boldsymbol{w}_g^r ) \| _{2}^{2} ]}\\
        % &-\tfrac{\eta _t}{2}\sum\nolimits_{\tau =1}^M{\mathbb{E} [ \| \sum\nolimits_{k=1}^K{\alpha _k^r \nabla F_k( \boldsymbol{w}_{k,\tau -1}^{{r-s_k^r}} ;D_{k,\tau}^{r-s_k^r} )} \| _{2}^{2} ]}\\
        % \le& \tfrac{\eta _t}{2}\sum\nolimits_{\tau =1}^M\mathbb{E} [ \| \sum\nolimits_{k=1}^K{\alpha _k^r \nabla F_k( \boldsymbol{w}_{k,\tau -1}^{{r-s_k^r}} ;D_{k,\tau}^{r-s_k^r} )}-\\
        % &\nabla F( \boldsymbol{w}_g^r ) \| _{2}^{2} ]-\tfrac{\eta _t}{2}\sum\nolimits_{\tau =1}^M{\mathbb{E} [ \| \nabla F( \boldsymbol{w}_g^r ) \| _{2}^{2} ]}\\
        \mathop {\le}^{(a)}&-\tfrac{\eta _tM}{2}\mathbb{E} [ \| \nabla F( \boldsymbol{w}_g^r ) \| _{2}^{2} ] + \eta _t\sum\nolimits_{k=1}^K{\alpha _k^r}\sum\nolimits_{\tau =1}^M \\
        &\mathbb{E} [ \| \nabla F_k( \boldsymbol{w}_g^{r-s_k^r} )-\nabla F_k( {\boldsymbol{w}_k}^{{r-s_k^r},\tau -1} ) \| _{2}^{2} ]\\
        &+{\eta _t\sum\nolimits_{\tau =1}^M{\mathbb{E} [ \| \nabla F(\boldsymbol{w}_g^r ) -\sum\nolimits_{k=1}^K{\alpha _k^r \nabla F_k( \boldsymbol{w}_g^{r-s_k^r} )} \| _{2}^{2} ]}}\\
        \mathop {\le}^{(b)}&-\tfrac{\eta _tM}{2}\mathbb{E} [ \| \nabla F( \boldsymbol{w}_g^r ) \| _{2}^{2} ] \\
        &+\eta _tL^2\sum\nolimits_{k=1}^K{\alpha _k^r}\sum\nolimits_{\tau =1}^M{\mathbb{E} [ \| \boldsymbol{w}_g^{r-s_k^r} -\boldsymbol{w}_{k,\tau -1}^{{r-s_k^r}} \| _{2}^{2} ]}\\
        &+\underset{C_1}{\underbrace{\eta _t\sum\nolimits_{\tau =1}^M{\mathbb{E} [ \| \nabla F(\boldsymbol{w}_g^r ) -\sum\nolimits_{k=1}^K{\alpha _k^r \nabla F_k( \boldsymbol{w}_g^{r-s_k^r} )} \| _{2}^{2} ]}}}
        % \underset{C_2}{\underbrace{\eta _t\sum\nolimits_{\tau =1}^M{\mathbb{E} [ \| \sum\nolimits_{k=1}^K{\alpha _k^r ( \nabla F_k( \boldsymbol{w}_g^{r-s_k^r} ) -\nabla F_k( {\boldsymbol{w}_k}^{{r-s_k^r},\tau -1} ) )} \| _{2}^{2} ]}}}
    \end{aligned}
\end{equation}
where (a) follows the inequality \eqref{sum square} and Jensen's inequality, and (b) follows the \eqref{ass1-1} in assumption 1 and $\boldsymbol{w}_g^{r-s_k^r}=\boldsymbol{w}_k^{r-s_k^r,0}$.

For \eqref{prooftheorem1B1}, we further need to bound the term $C_1$ as follow,
\begin{align}\label{prooftheorem1C1}
    &C_1\notag\\
    =&\eta _t\sum\nolimits_{\tau =1}^M{\mathbb{E} [ \| \nabla F( \boldsymbol{w}_g^r ) -\sum\nolimits_{k=1}^K{\alpha _k^r \nabla F_k( \boldsymbol{w}_g^{r-s_k^r} )} \| _{2}^{2} ]}\notag\\
    =&\eta _t\sum\nolimits_{\tau =1}^M{\mathbb{E} [ \| \sum\nolimits_{k=1}^K{\alpha _k^r ( \nabla F( \boldsymbol{w}_g^r ) -\nabla F_k( \boldsymbol{w}_g^{r-s_k^r} ) )} \| _{2}^{2} ]}\notag\\
    \mathop {\le}^{(a)}&\eta _t\sum\nolimits_{k=1}^K{\alpha _k^r \sum\nolimits_{\tau =1}^M{\mathbb{E} [ \| \nabla F( \boldsymbol{w}_g^r ) -\nabla F_k( \boldsymbol{w}_g^{r-s_k^r} ) \| _{2}^{2} ]}}\notag\\
    =&\eta _t\sum\nolimits_{k=1}^K\alpha _k^r \sum\nolimits_{\tau =1}^M \mathbb{E} [ \| \nabla F( \boldsymbol{w}_g^r ) -\nabla F( \boldsymbol{w}_g^{r-s_k^r} )\notag\\ 
    & +\nabla F( \boldsymbol{w}_g^{r-s_k^r} ) -\nabla F_k( \boldsymbol{w}_g^{r-s_k^r} ) \| _{2}^{2} ]\notag\\
    \mathop {\le}^{(b)}&\eta _t\sum\nolimits_{k=1}^K{\alpha _k^r \sum\nolimits_{\tau =1}^M{2\mathbb{E} [ \| \nabla F( \boldsymbol{w}_g^r ) -\nabla F( \boldsymbol{w}_g^{r-s_k^r} ) \| _{2}^{2} ]}}+\notag\\
    &\eta _t\sum\nolimits_{k=1}^K{\alpha _k^r \sum\nolimits_{\tau =1}^M{2\mathbb{E} [ \| \nabla F( \boldsymbol{w}_g^{r-s_k^r} ) -\nabla F_k( \boldsymbol{w}_g^{r-s_k^r} ) \| _{2}^{2} ]}}\notag\\
    \mathop {\le}^{(c)}&2L\eta _t\sum\nolimits_{k=1}^K{\alpha _k^r \sum\nolimits_{\tau =1}^M{\mathbb{E} [ \| \boldsymbol{w}_g^r -\boldsymbol{w}_g^{r-s_k^r} \| _{2}^{2} ]}}+\notag\\
    &2\eta _t\sum\nolimits_{k=1}^K{\alpha _k^r \sum\nolimits_{\tau =1}^M{\mathbb{E} [ \| \nabla F( \boldsymbol{w}_g^{r-s_k^r} ) -\nabla F_k( \boldsymbol{w}_g^{r-s_k^r} ) \| _{2}^{2} ]}}\notag\\
    \mathop {\le}^{(d)}&2\eta _tML^2\varepsilon ^2+2\eta _tM\zeta 
\end{align}
where (a) follows the Jensen's Inequality, (b) follows the inequality $\| \boldsymbol{x}_1+\boldsymbol{x}_2 \| _{2}^{2}\le 2\| \boldsymbol{x}_1 \| _{2}^{2}+2\| \boldsymbol{x}_2 \| _{2}^{2}$, (c) follows the \eqref{ass1-1} in assumption 1, and (d) follows the \eqref{ass2-1} in Assumption 2 and the \eqref{ass3-2} in Assumption 3.
% \begin{equation}
%     \begin{aligned}
%         C_2=&\eta _t\sum\nolimits_{\tau =1}^M \mathbb{E} [ \| \sum\nolimits_{k=1}^K \alpha _k^r ( \nabla F_k( \boldsymbol{w}_g^{r-s_k^r} ) \\
%         &-\nabla F_k( {\boldsymbol{w}_k}^{{r-s_k^r},\tau -1} ) )  \| _{2}^{2} ]\\
%         \mathop {\le}^{(a)}&\eta _t\sum\nolimits_{k=1}^K{\alpha _k^r}\sum\nolimits_{\tau =1}^M \mathbb{E} [ \| \nabla F_k( \boldsymbol{w}_g^{r-s_k^r} )\\
%         & -\nabla F_k( {\boldsymbol{w}_k}^{{r-s_k^r},\tau -1} ) \| _{2}^{2} ]\\
%         \mathop {\le}^{(b)}&\eta _tL^2\sum\nolimits_{k=1}^K{\alpha _k^r}\sum\nolimits_{\tau =1}^M{\mathbb{E} [ \| \boldsymbol{w}_g^{r-s_k^r} -\boldsymbol{w}_{k,\tau -1}^{{r-s_k^r}} \| _{2}^{2} ]}
%     \end{aligned}
% \end{equation}

Combining \eqref{prooftheorem1A1}-\eqref{prooftheorem1C1}, we can obtain
\begin{equation}\label{prooftheorem1-A_1}
    \begin{aligned}
        A_1\le& B_1+B_2+B_3\\
        % \le& \delta \mathbb{E} [ \| \nabla F( \boldsymbol{w}_g^r ) \| _{2}^{2} ] -\tfrac{\eta _tM}{2}\mathbb{E} [ \| \nabla F( \boldsymbol{w}_g^r ) \| _{2}^{2} ] +C_1\\
        % &+\eta _tL^2\sum\nolimits_{k=1}^K{\alpha _k^r}\sum\nolimits_{\tau =1}^M{\mathbb{E} [ \| \boldsymbol{w}_g^{r-s_k} -\boldsymbol{w}_{k}^{{r-s_k},\tau -1}\| _{2}^{2} ]}\\
        \le& ( \delta -\tfrac{\eta _tM}{2} ) \mathbb{E} [ \| \nabla F( \boldsymbol{w}_g^r ) \| _{2}^{2} ] +2\eta _tML^2\varepsilon ^2+2\eta _tM\zeta \\
        & +\eta _tL^2\sum\nolimits_{k=1}^K{\alpha _k^r}\sum\nolimits_{\tau =1}^M{\mathbb{E} [ \| \boldsymbol{w}_g^{r-s_k} -\boldsymbol{w}_{k,\tau -1}^{{r-s_k}}\| _{2}^{2} ]}
    \end{aligned}
\end{equation}

\subsection{Bound of $A_2$}
In this section, we derive the bound of $A_2$,
\begin{equation}\label{prooftheorem1A2}
    \begin{aligned}
        A_2=&\mathbb{E} [ \| \boldsymbol{w}_g^{r+1} -\boldsymbol{w}_g^{r} \| _{2}^{2} ]\\ =&\mathbb{E} [ \| \tilde{\boldsymbol{w}}^{r} -\boldsymbol{w}_g^{r} +\sum\nolimits_{k=1}^K{\alpha _k^{r} \Delta \boldsymbol{w}_k^{r}}+\tilde{\boldsymbol{n}}^{r} \| _{2}^{2} ] \\
        \mathop {\le}^{(a)}&2\mathbb{E} [ \| \tilde{\boldsymbol{w}}^{r} -\boldsymbol{w}_g^{r} \| _{2}^{2} ] +4\mathbb{E} [ \| \sum\nolimits_{k=1}^K{\alpha _k^{r} \Delta \boldsymbol{w}_k^{r}} \| _{2}^{2} ] \\
        &+4\mathbb{E} [ \| \tilde{\boldsymbol{n}}^{r} \| _{2}^{2} ] \\
        \mathop {=}^{(b)}&2\underset{B_1}{\underbrace{\mathbb{E} [ \| \tilde{\boldsymbol{w}}^{r} -\boldsymbol{w}_g^{r} \| _{2}^{2} ] }}+4\underset{B_2}{\underbrace{\mathbb{E} [ \| \sum\nolimits_{k=1}^K{\alpha _k^{r} \Delta \boldsymbol{w}_k^{r}} \| _{2}^{2} ] }}\\
        &+\frac{4d\sigma _{u}^{2}}{( \sum\nolimits_{k\in K}^{}{a_k^rp_k^{r}} ) ^2}
    \end{aligned}
\end{equation}
where (a) follows the inequality \eqref{sum square}, and (b) follows noise mean square calculation.

We analyze $B_1$ and $B_2$ separately. We bound the term $B_1$ as ,
\begin{equation}
    \begin{aligned}
        B_1=&\mathbb{E} [ \| \tilde{\boldsymbol{w}}^{r} -\boldsymbol{w}_g^{r} \| _{2}^{2} ] \\
        =&\mathbb{E} [ \| \sum\nolimits_{k=1}^K{\alpha _k^{r} ( \boldsymbol{w}_g^{r-s_k^r} -\boldsymbol{w}_g^{r} )} \| _{2}^{2} ] \\
        =&K\sum\nolimits_{k=1}^K{{{(\alpha _k^{r})}^2} \mathbb{E} [ \| \boldsymbol{w}_g^{r-s_k^r} -\boldsymbol{w}_g^{r} \| _{2}^{2} ]}\\
        \le& \varepsilon ^2K\sum\nolimits_{k=1}^K{{(\alpha _k^{r})}^2}
    \end{aligned}
\end{equation}
we study the term $B_2$ as follow, 
\begin{align}\label{prooftheorem1B22}
    &B_2\notag\\
    =&\mathbb{E} [ \| \sum\nolimits_{k=1}^K{\alpha _k^r \Delta \boldsymbol{w}_k^r} \| _{2}^{2} ] \notag\\
    \mathop {\le}^{(a)}&\sum\nolimits_{k=1}^K{\alpha _k^r \mathbb{E} [ \| \Delta \boldsymbol{w}_k^r \| _{2}^{2} ]}\notag\\
    % =&\sum\nolimits_{k=1}^K\alpha _k^r \mathbb{E} [ \| -\eta _t\sum\nolimits_{\tau =1}^M\nabla F_k( \boldsymbol{w}_{k,\tau-1}^{r-s_k^r} ;\\
    % &D_{k,\tau}^{r-s_k^r} ) \| _{2}^{2} ]\\
    =&{\eta _t}^2\sum\nolimits_{k=1}^K\alpha _k^r \mathbb{E} [ \| \sum\nolimits_{\tau =1}^M\nabla F_k( \boldsymbol{w}_{k,\tau-1}^{r-s_k^r} ;D_{k,\tau}^{r-s_k^r} ) \| _{2}^{2} ]\notag\\
    \le& {\eta _t}^2M\sum\nolimits_{k=1}^K{\alpha _k^r \sum\nolimits_{\tau =1}^M{\mathbb{E} [ \| \nabla F_k( \boldsymbol{w}_{k,\tau-1}^{r-s_k^r} ;D_{k,\tau}^{r-s_k^r} ) \| _{2}^{2} ]}}\notag\\
    \mathop {=}^{(b)}&{\eta _t}^2M\sum\nolimits_{k=1}^K\alpha _k^r \sum\nolimits_{\tau =1}^M\mathbb{E} [ \| \nabla F_k( \boldsymbol{w}_{k,\tau-1}^{r-s_k^r} ;D_{k,\tau}^{r-s_k^r} ) \notag\\
    &-\nabla F_k( \boldsymbol{w}_{k,\tau-1}^{r-s_k^r} ) \| _{2}^{2} ]\notag\\
    &+{\eta _t}^2M\sum\nolimits_{k=1}^K{\alpha _k^r \sum\nolimits_{\tau =1}^M{\mathbb{E} [ \| \nabla F_k( \boldsymbol{w}_{k,\tau-1}^{r-s_k^r} ) \| _{2}^{2} ]}}\notag\\
    \mathop {\le}^{(c)}&{\eta _t}^2M^2\sigma ^2\notag\\
    &+{\eta _t}^2M\sum\nolimits_{k=1}^K{\alpha _k^r \sum\nolimits_{\tau =1}^M{\mathbb{E} [ \| \nabla F_k( \boldsymbol{w}_{k,\tau-1}^{r-s_k^r} ) \| _{2}^{2} ]}}\notag\\
    \mathop {=}^{(d)}&{\eta _t}^2M^2\sigma ^2+{\eta _t}^2M\sum\nolimits_{k=1}^K\alpha _k^r \sum\nolimits_{\tau =1}^M \mathbb{E} [ \| \nabla F_k( \boldsymbol{w}_{k,\tau-1}^{r-s_k^r} )\notag\\
    & -\nabla F_k( \boldsymbol{w}_g^{r-s_k^r} ) +\nabla F_k( \boldsymbol{w}_g^{r-s_k^r} ) \| _{2}^{2} ]\notag\\
    \mathop {\le}^{(e)}&{\eta _t}^2M^2\sigma ^2+2{\eta _t}^2M^2\sum\nolimits_{k=1}^K{\alpha _k^r \mathbb{E} [ \| \nabla F_k( \boldsymbol{w}_g^{r-s_k^r} ) \| _{2}^{2} ]}\notag\\
    &+2{M\eta _t}^2\sum\nolimits_{k=1}^K\alpha _k^r \sum\nolimits_{\tau =1}^M\mathbb{E} [ \| \nabla F_k( \boldsymbol{w}_{k,\tau-1}^{r-s_k^r} ) \notag\\
    &-\nabla F_k( \boldsymbol{w}_g^{r-s_k^r} ) \| _{2}^{2} ]\notag\\
    \mathop {\le}^{(f)}&{\eta _t}^2M^2\sigma ^2\notag\\
    &+2{M\eta _t}^2L^2\sum\nolimits_{k=1}^K{\alpha _k^r \sum\nolimits_{\tau =1}^M{\mathbb{E} [ \| \boldsymbol{w}_{k,\tau-1}^{r-s_k^r} -\boldsymbol{w}_g^{r-s_k^r} \| _{2}^{2} ]}}\notag\\
    & +2{\eta _t}^2M^2\sum\nolimits_{k=1}^K\alpha _k^r \mathbb{E} [ \| \nabla F_k( \boldsymbol{w}_g^{r-s_k^r} ) -\nabla F( \boldsymbol{w}_g^{r-s_k^r} )\notag\\
    & +\nabla F( \boldsymbol{w}_g^{r-s_k^r} ) \| _{2}^{2} ]\notag\\
    % \mathop {\le}^{(g)}&{\eta _t}^2M^2\sigma ^2\notag\\
    % &+2{M\eta _t}^2L^2\sum\nolimits_{k=1}^K{\alpha _k^r \sum\nolimits_{\tau =1}^M{\mathbb{E} [ \| \boldsymbol{w}_{k,\tau-1}^{r-s_k^r} -\boldsymbol{w}_g^{r-s_k^r} \| _{2}^{2} ]}}\notag\\
    % &+4{\eta _t}^2M^2\sum\nolimits_{k=1}^K{\alpha _k^r \mathbb{E} [ \| \nabla F_k( \boldsymbol{w}_g^{r-s_k^r} ) -\nabla F( \boldsymbol{w}_g^{r-s_k^r} ) \| _{2}^{2} ]}\notag\\
    % &+4{\eta _t}^2M^2\sum\nolimits_{k=1}^K{\alpha _k^r \mathbb{E} [ \| \nabla F( \boldsymbol{w}_g^{r-s_k^r} ) \| _{2}^{2} ]}\notag\\
    \mathop {\le}^{(g)}&{\eta _t}^2M^2\sigma ^2+4{\eta _t}^2M^2\zeta +4{\eta _t}^2M^2\beta ^2\mathbb{E} [ \| \nabla F( \boldsymbol{w}_g^r ) \| _{2}^{2} ] \notag\\
    &+2{M\eta _t}^2L^2\sum\nolimits_{k=1}^K{\alpha _k^r \sum\nolimits_{\tau =1}^M{\mathbb{E} [ \| \boldsymbol{w}_{k,\tau-1}^{r-s_k^r} -\boldsymbol{w}_g^{r-s_k^r} \| _{2}^{2} ]}}
\end{align}
where (a) follows Jensen's Inequality, (b) and (d) follow equality \eqref{square expectation decomposition}, (c) follows the equality \eqref{gradient variance}, (e) follow equality \eqref{sum square}, (f) follows the \eqref{ass1-2} in Assumption 1, and (g) follows the \eqref{ass2-1} in Assumption 2 and \eqref{ass3-3} in Assumption 3.

Combining \eqref{prooftheorem1A2}-\eqref{prooftheorem1B22}, we can obtain
\begin{equation}\label{prooftheorem1-A_2}
    \begin{aligned}
        A_2\le& 2B_1+4B_2+\frac{4d\sigma _{u}^{2}}{( \sum\nolimits_{k\in K}^{}{a_k^rp_k^r} ) ^2}\\
        \le& \frac{4d\sigma _{u}^{2}}{( \sum\nolimits_{k\in K}^{}{a_k^rp_k^r} ) ^2} +4{\eta _t}^2M^2\sigma ^2+16{\eta _t}^2M^2\zeta \\
        &+16{\eta _t}^2M^2\beta ^2\mathbb{E} [ \| \nabla F( \boldsymbol{w}_g^r ) \| _{2}^{2} ]+2\varepsilon ^2K\sum\nolimits_{k=1}^K{({\alpha _k^r})^2}\\
        &+8{M\eta _t}^2L^2\sum\nolimits_{k=1}^K\alpha _k^r \sum\nolimits_{\tau =1}^M\mathbb{E} [ \| \boldsymbol{w}_{k,{\tau -1}}^{r-s_k^r}\\
        & -\boldsymbol{w}_g^{r-s_k^r} \| _{2}^{2} ]
    \end{aligned}
\end{equation}

\subsection{Proof of Theorem 1}
Combining \eqref{prooftheorem1-2}, \eqref{prooftheorem1-A_1} and \eqref{prooftheorem1-A_2}, we obtain
\begin{equation}\label{prooftheorem1-c}
    \begin{aligned}
        &\mathbb{E} [ F( \boldsymbol{w}_g^{r+1} ) ] \\
        \le &\mathbb{E} [ F( \boldsymbol{w}_g^r ) ] +A_1+\frac{L}{2}A_2\\
        \mathop {\le}^{(a)}&\mathbb{E} [ F( \boldsymbol{w}_g^r ) ] +( \delta -\tfrac{\eta _tM}{2}+8L\eta _{t}^{2}M^2\beta ^2\\
        &+( \eta _tL^2+4M\eta _{t}^{2}L^3 ) \frac{4\eta _{t}^{2}M^3\beta ^2}{1-2\eta _{t}^{2}M^2} ) \mathbb{E} [ \| \nabla F( \boldsymbol{w}_g^r ) \| _{2}^{2} ] \\
        &+2\eta _tML^2\varepsilon ^2+2\eta _tM\zeta +L\varepsilon ^2K\sum\nolimits_{k=1}^K{({\alpha _k^r})^2}\\
        &+2L\eta _{t}^{2}M^2\sigma ^2+8L\eta _{t}^{2}M^2\zeta \\
        &+( \eta _tL^2+4M\eta _{t}^{2}L^3 ) \frac{\eta _{t}^{2}M^3\sigma ^2+4\eta _{t}^{2}M^3L^2\zeta}{1-2\eta _{t}^{2}M^2L^2}\\
        &+\frac{2Ld\sigma _{u}^{2}}{( \sum\nolimits_{k\in K}^{}{a_k^rp_k^r} ) ^2}
    \end{aligned}
\end{equation}
where (a) follows \eqref{lemma1} in Lemma 1.

By subtracting $F^*$ at both sides of \eqref{prooftheorem1-c}, we have
% \begin{align}
%     &\mathbb{E} [ F( \boldsymbol{w}_g^{r+1} ) -F^* ] \notag\\
%     % \le& \mathbb{E} [ F( \boldsymbol{w}_g^r ) -F^* ] +\mathbb{E} [ \| \nabla F( \boldsymbol{w}_g^r ) \| _{2}^{2} ]\cdot\notag\\
%     % &( \delta -\tfrac{\eta _tM}{2}+8L\eta _{t}^{2}M^2\beta ^2+( \eta _tL^2+4M\eta _{t}^{2}L^3 ) \frac{4\eta _{t}^{2}M^3\beta ^2}{1-2\eta _{t}^{2}M^2} )  \notag\\
%     % & +2\eta _tML^2\varepsilon ^2+2\eta _tM\zeta +L\varepsilon ^2K\sum\nolimits_{k=1}^K{({\alpha _k^r})^2}+2L\eta _{t}^{2}M^2\sigma ^2 \notag\\
%     % & +8L\eta _{t}^{2}M^2\zeta+( \eta _tL^2+4M\eta _{t}^{2}L^3 ) \frac{{\eta _t}^2M^3\sigma ^2+4\eta _{t}^{2}M^3L^2\zeta}{1-2\eta _{t}^{2}M^2L^2}\notag\\
%     % &+\frac{2Ld\sigma _{u}^{2}}{( \sum\nolimits_{k\in K}^{}{a_k^rp_k^r} ) ^2}\notag\\
%     % \mathop {\le}^{(a)}&( 1+2L\delta -L\eta _tM+8L^2\eta _{t}^{2}M\beta ^2\notag\\
%     % &+( \eta _tL^2+4M\eta _{t}^{2}L^3 ) \frac{8L\eta _{t}^{2}M^3\beta ^2}{1-2\eta _{t}^{2}M^2L^2} ) \mathbb{E} [ F( \boldsymbol{w}_g^r ) -F^* ] \notag\\
%     % &+2\eta _tML^2\varepsilon ^2+2\eta _tM\zeta +L\varepsilon ^2K\sum\nolimits_{k=1}^K{({\alpha _k^r})^2}\notag\\
%     % &+2L\eta _{t}^{2}M^2\sigma ^2+8L\eta _{t}^{2}M^2\zeta \notag\\
%     % &+( \eta _tL^2+4M\eta _{t}^{2}L^3 ) \frac{{\eta _t}^2M^3\sigma ^2+4{\eta _t}^2M^3L^2\zeta}{1-2\eta _{t}^{2}M^2L^2}\notag\\
%     % &+\frac{2Ld\sigma _{u}^{2}}{( \sum\nolimits_{k\in K}^{}{a_k^rp_k^r} ) ^2}\notag\\
%     \mathop {\le}^{(a)} & A^r \mathbb{E} [ F( \boldsymbol{w}_g^r ) -F^* ] +G^r
% \end{align}
\begin{equation}
    \mathbb{E} [ F( \boldsymbol{w}_g^{r+1} ) -F^* ]\mathop {\le}^{(a)} A^r \mathbb{E} [ F( \boldsymbol{w}_g^r ) -F^* ] +G^r
\end{equation}
with
\begin{equation}
\begin{aligned}
    A^r = &1+2L\delta -L\eta M+8L^2\eta^{2}M\beta ^2\\
    &+( \eta L^2+4M\eta^{2}L^3 ) \frac{8L\eta^{2}M^3\beta ^2}{1-2\eta^{2}M^2L^2},
\end{aligned}
\end{equation}
{and}
\begin{equation}
\begin{aligned}
    G^r =&\underbrace{(2\eta M+8L\eta M^2+\frac{4\eta^2M^3L^2(\eta L^2+4M\eta^2L^3)}{1-2\eta^2M^2L^2})\zeta}_{(a)}\\
    & + \underbrace{2\eta ML^2\epsilon^2}_{(b)}+\underbrace{(2\eta^2 LM^2+\frac{(\eta L^2+4M\eta^2L^3)\eta^2M^3}{1-2\eta^2M^2L^2})\sigma^2}_{(c)}\\
    &+ \underbrace{L\epsilon^2K\sum\nolimits_{k=1}^K(\alpha_k^r)^2}_{(d)}+ \underbrace{\frac{2Ld\sigma_n^2}{(\sum\nolimits_{k=1}^K b_k^r p_k^r)^2}}_{(e)},
\end{aligned}
\end{equation}
where (a) follows \eqref{eq:lemma2} in Lemma 2.

Assume the FL algorithm terminates after $R$ rounds, given an initial global model $\boldsymbol{w}_1$,we carry out recursions as
\begin{align}
    &\mathbb{E} [ F( \boldsymbol{w}_g^{R+1} ) -F^* ]\notag \\
    \le& A^R \mathbb{E} [ F( \boldsymbol{w}_g^R ) -F^* ] +G^R \notag \\
    \le& A^R A^{R-1} \mathbb{E} [ F( \boldsymbol{w}_g^{R-1} ) -F^* ] +G^{R-1} +G^R \notag \\
    \le& \cdots\notag  \\
    \le& \prod_{t=1}^{R}{A^r\mathbb{E} [ F( \boldsymbol{w}_g^1 ) -F^* ]}+\sum_{t=1}^{R}{( \prod_{i=r+1}^{R}{A^r} ) G^r +G^R}
\end{align}

Thus, this completes the proof.
}

% \input{draft/7_appendix(simplified).tex}

% \vspace{12pt}
% \color{red}
% IEEE conference templates contain guidance text for composing and formatting conference papers. Please ensure that all template text is removed from your conference paper prior to submission to the conference. Failure to remove the template text from your paper may result in your paper not being published.

\end{document}